\RequirePackage{fix-cm}
\documentclass[smallextended]{svjour3}       
\smartqed  

\usepackage{pgfplots,xifthen,bm,tikz,tikz-cd,adjustbox,afterpage,algorithmic}
\usetikzlibrary{calc, positioning}
\usepackage{array}
\usepackage{eqparbox}

\usepackage{etoolbox}  
\makeatletter
\patchcmd{\algorithmic}{\addtolength{\ALC@tlm}{\leftmargin} }{\addtolength{\ALC@tlm}{\leftmargin}}{}{}
\newcommand\footnoteref[1]{\protected@xdef\@thefnmark{\ref{#1}}\@footnotemark}
\makeatother
\usetikzlibrary{patterns}
\usepackage{pgfplots}
\usepackage{float}
\usepackage{graphicx}

\usepackage{caption}
\usepackage{subcaption}
\usepackage{multirow}
\usepackage{booktabs}

\usepackage{amsmath}

\usepackage{pgfplotstable}
\usepackage[ruled,vlined]{algorithm2e}
\usepackage{amsmath}
\usepackage{amsfonts}
\SetAlFnt{\footnotesize}
\SetKwInput{KwInit}{Initialization}
\usepackage{natbib}

\usepackage{hyperref}

\journalname{Machine Learning}
\begin{document}
	
	\title{Adaptive Deep Forest for Online Learning from Drifting Data Streams
	}
	
	\titlerunning{Adaptive Deep Forest for Online Learning from Drifting Data Streams}        
	
	\author{\L{}ukasz Korycki         \and
		Bartosz Krawczyk 
	}
	
	
	\institute{\L{}. Korycki and B. Krawczyk  \at
		Department of Computer Science, Virginia Commonwealth University, Richmond VA, USA \\
		\email{\{koryckil,bkrawczyk\}@vcu.edu}           \\
	}
	
	\date{Received: date / Accepted: date}

	\maketitle
	
	\begin{abstract}
		Learning from data streams is among the most vital fields of contemporary data mining. The online analysis of information coming from those potentially unbounded data sources allows for designing reactive up-to-date models capable of adjusting themselves to continuous flows of data. While a plethora of shallow methods have been proposed for simpler low-dimensional streaming problems, almost none of them addressed the issue of learning from complex contextual data, such as images or texts. The former is represented mainly by adaptive decision trees that have been proven to be very efficient in streaming scenarios. The latter has been predominantly addressed by offline deep learning. In this work, we attempt to bridge the gap between these two worlds and propose Adaptive Deep Forest (ADF) -- a natural combination of the successful tree-based streaming classifiers with deep forest, which represents an interesting alternative idea for learning from contextual data. The conducted experiments show that the deep forest approach can be effectively transformed into an online algorithm, forming a model that outperforms all state-of-the-art shallow adaptive classifiers, especially for high-dimensional complex streams.
		
	\end{abstract}
	
	\section{Introduction}
	\label{sec:int}
	
	Online machine learning generalizes the standard offline approach by overcoming the naive assumption that a single batch of data may sufficiently represent a given problem forever. This is simply not true as most of the real-world phenomena are naturally dynamic \citep{Ditzler:2015,Tickle:2019} and we need to understand what may cause them to change \cite{Arrieta:2020}. Nothing should explain it better than the most recent crisis caused by the coronavirus pandemic, which dramatically affected the everyday lives of many people. As a result, many companies encountered a sudden shift in the characteristics of their potential clients, which invalidated their models, meticulously built upon representative yet now outdated collections of data. 
	
	Although the current crisis is extraordinarily severe, changes on a smaller scale happens much more frequently in a wide range of domains \citep{Gama:2014}. Furthermore, even if our data is stationary, in many cases, we may still be unable to collect a complete representation of our problem at once, requiring some time for incremental aggregation of knowledge. Recognizing the problem is the first step towards developing more resilient, accurate and flexible real-time systems, which will provide online convergence to offline counterparts and, in addition, reactivity in case of changes.
	
	In lifelong data stream mining, the assumption that we should stay connected with our data sources is the fundamental rule upon which all modeling concepts are constructed \citep{Gama:2012}. They are focused on aggregating up-to-date knowledge and forgetting outdated concepts \citep{Ditzler:2015}. Most importantly, every streaming algorithm should be capable of reacting to a phenomenon known as concept drift, which we define as a change in class-conditional distributions \citep{Gama:2014}. Several different approaches to this task have been proposed over many years \citep{Krawczyk:2017, Lu:2019}. However, even if the algorithms are getting more and more effective, most of them focus only on shallow learning and do not address the problem of deep learning from contextual data \citep{Sahoo:2018}. 
	
	\smallskip
	\noindent \textbf{Goal.} To develop a novel deep learning architecture capable of efficient learning from high-dimensional complex data streams, while being capable of handling concept drift.
	
	\smallskip
	\noindent \textbf{Motivation.} Contemporary problems generate continuously streaming, drifting and, at the same time, complex, high-dimensional data. Streams of images, text, or other tensor-based data are becoming more and more frequent in the modern world \citep{Liu:2020}. However, there exists a gap between such data and our capabilities of processing it. Data stream mining domain focuses on such aspects as adaptation to changes, fast reactivity, robustness to concept drift, or learning from massive data sets on-the-fly \citep{Korycki:2019,Ramirez-Gallego:2017}. However, all of those solutions are designed primarily for shallow data and lack efficient ways of dealing with high-dimensional problems \citep{Ryan:2019}. At the same time, deep learning offers powerful tools for embedding such problems and extracting features from them with powerful generalization capabilities. However, this set-up is predominantly limited to the offline setting and no adaptation to changes and drifts is being considered \citep{Fayek:2020}. Therefore, there is a need to bridge those two worlds by developing novel hybrid solutions that combine the mechanisms and solutions from those domains in order to create truly adaptive deep learning systems.  
	
	\smallskip
	\noindent \textbf{Overview.} In this work, we present the Adaptive Deep Forest algorithm, which incorporates very effective streaming decision trees into a deep forest architecture \citep{Zhou:2017df}. By doing so, we aim at developing a deep online model capable of learning from drifting high-dimensional streams of images, texts and time series. We combine a deep architecture based on several cascade layers of forest ensembles with multi-grained representations of input high-dimensional data. Each cascade layer is trained using an adaptive ensemble algorithm, allowing us to combine the capabilities of embedding high-dimensional data with excellent reactivity to concept drift.
	
	\smallskip
	\noindent \textbf{Main contributions.} This paper offers the following advances towards creating efficient deep learning algorithms for drifting data streams.
	
	\begin{itemize}
		\smallskip
		\item \textbf{Deep learning architecture for data streams.} We propose Adaptive Deep Forest, a novel deep architecture capable of learning from drifting and complex data streams. ADF is based on several cascading layers, each realized as ensembles of adaptive decision trees and supplied with additional information coming from extracted features at varying resolutions. This way we are able to combine the advantage of tree-based architectures (robustness to concept drift) with deep architectures (handling complex data and powerful predictive power).
		
		\smallskip
		\item \textbf{Online multi-grain scanning.} We use a sliding window of varying resolutions to extract features coming from different scales and use them as an additional input for each cascade layer of ADF. This allows for efficient handling of complex and high-dimensional data, embedding its information into the ADF layers. With such a multi--view approach, we are able to extract diverse features that improve the learning capabilities of ADF and allow for creating more specialized and accurate adaptive ensemble classifiers.
		
		\smallskip
		\item \textbf{Handling concept drift.} By using ensembles of adaptive decision trees in each layer we are able to handle concept drifts occurring at global (entire stream), as well as local (specific sub-regions) levels. Combining information coming from multi-grain scanning with layer-based drift detection allows ADF to efficiently adapt to changes, without compromising its learning speed, while avoiding a need for rebuilding its deep architecture. 
		
		\smallskip
		\item \textbf{Extensive experimental study.} We evaluate the efficacy of ADF against 13 state--of--the--art algorithms dedicated to data stream mining, using a wide collection of popular shallow and deep benchmarks. We perform not only comparative analysis, but also an in-depth analysis of the ADF learning procedure and properties. This allows us to gain unique insights into the propagation of knowledge between the ADF layers, their impact on the ADF performance, as well as into the ADF reactivity to various types of concept drift. 
	\end{itemize}
	
	\section{Learning from streams}
	\label{sec:stream}
	
	\noindent \textbf{Data streams.} This domain deals with problems in which data arrives continuously from a given source. A streaming classifier should be able to adequately incorporate the knowledge encoded in the new instances. Data stream is defined as a sequence ${<S_1, S_2, ..., S_n,...>}$, where each element $S_j$ is a new instance. In this paper, we assume the supervised learning scenario with classification task and thus we define each instance as $S_j \sim p_j(x^1,\cdots,x^d,y) = p_j(\mathbf{x},y)$, where $p_j(\mathbf{x},y)$ is a joint distribution of the $j$-th instance, defined by a $d$-dimensional feature space and assigned to class $y$. Each instance is independent and drawn randomly from a probability distribution $\Psi_j (\mathbf{x},y)$. 
	
	\smallskip
	\noindent \textbf{Concept drift.} When all instances come from the same distribution, we deal with a stationary data stream. In real-world applications data very rarely falls under stationary assumptions \citep{Masegosa:2020}. It is more likely to evolve over time and form temporary concepts, being subject to concept drift \citep{Lu:2019}. This phenomenon affects various aspects of a data stream and thus can be analyzed from multiple perspectives \citep{Pinage:2020}. One cannot simply claim that a stream is subject to the drift. It needs to be analyzed and understood in order to be handled adequately to specific changes that occur \citep{Goldenberg:2019,Goldenberg:2020}. More precise approaches may help us achieving faster and more accurate adaptation \citep{Shaker:2015}. One must also be able to distinguish drift from noise \citep{RCano:2019}, as well as recognize if the drift is not an adversarial one injected via a poisoning attack \citep{Korycki:2020acd}. Let us now discuss the major aspects of concept drift and its characteristics. 
	
	Firstly, we need to take into account how concept drift impacts the learned decision boundaries, distinguishing between real and virtual concept drifts \citep{Oliveira:2019}. The former influences previously learned decision rules or classification boundaries, decreasing their relevance for newly incoming instances. Real drift affects posterior probabilities $p_j(y|\mathbf{x})$ and additionally may impact unconditional probability density functions. It must be tackled as soon as it appears since it impacts negatively the underlying classifier. Virtual concept drift affects only the distribution of features $\mathbf{x}$ over time:
	
	\begin{equation}
	\widehat{p}_j(\mathbf{x}) = \sum_{y \in Y} p_j(\mathbf{x},y),
	\label{eq:cd2}
	\end{equation}
	
	\noindent where $Y$ is a set of possible values taken by $S_j$. While it seems less dangerous than real concept drift, it cannot be ignored. Despite the fact that only the values of features change, it may trigger false alarms and thus force unnecessary and costly adaptations. 
	
	It is important to distinguish between global and local concept drifts~\citep{Gama:2006}. The former one affects the entire stream, while the latter one affects only certain parts of it (e.g., regions of the feature space, individual clusters of instances, or subsets of classes). Determining the locality of changes is of high importance, as rebuilding the entire classification model may not be necessary. Instead, one may update only certain parts of the model or sub-models, leading to a more efficient adaptation.
	
	One must also take into account the speed of changes. Here we distinguish between sudden, gradual, and incremental concept drifts~\citep{Lu:2019}. 
	
	\begin{itemize}
		\item \textbf{Sudden concept drift} is a case when instance distribution abruptly changes with $t$-th example arriving from the stream:
		
		\begin{equation}
		p_j(\mathbf{x},y) =
		\begin{cases}
		D_0 (\mathbf{x},y),       & \quad \text{if } j < t\\
		D_1 (\mathbf{x},y),  & \quad \text{if } j \geq t.
		\end{cases}
		\label{eq:cd3}
		\end{equation}
		
		\smallskip
		\item \textbf{Incremental concept drift} is a case when we have a continuous progression from one concept to another (thus consisting of multiple intermediate concepts in between), such that the distance from the old concept is increasing, while the distance to the new concept is decreasing:
		
		\begin{equation}
		p_j(\mathbf{x},y) =
		\begin{cases}
		D_0 (\mathbf{x},y),       &  \text{if } j < t_1\\
		(1 - \alpha_j) D_0 (\mathbf{x},y) + \alpha_j D_1 (\mathbf{x},y),       &\text{if } t_1 \leq j < t_2\\
		D_1 (\mathbf{x},y),  &  \text{if } t_2 \leq j
		\end{cases}
		\label{eq:cd4}
		\end{equation}
		
		\noindent where
		
		\begin{equation}
		\alpha_j = \frac{j - t_1}{t_2 - t_1}.
		\label{eq:cd5}
		\end{equation}
		
		\smallskip
		\item \textbf{Gradual concept drift} is a case where instances arriving from the stream oscillate between two distributions during the duration of the drift, with the old concept appearing with decreasing frequency:
		
		\begin{equation}
		p_j(\mathbf{x},y) =
		\begin{cases}
		D_0 (\mathbf{x},y),       &  \text{if } j < t_1\\
		D_0 (\mathbf{x},y),       &  \text{if } t_1 \leq j < t_2 \wedge \delta > \alpha_j\\
		D_1 (\mathbf{x},y),       &  \text{if } t_1 \leq j < t_2 \wedge \delta \leq \alpha_j\\
		D_1 (\mathbf{x},y),  &  \text{if } t_2 \leq j,
		\end{cases}
		\label{eq:cd4}
		\end{equation}
		\noindent where $\delta \in [0,1]$ is a random variable. 
	\end{itemize}
	
	\noindent Finally, in many scenarios it is possible that a previously seen concept from $k$-th iteration may reappear D$_{j+1}$ = D$_{j-k}$ over time \citep{Sobolewski:2017}. One may store models specialized in previously seen concepts in order to speed up recovery rates after a known concept re-emerges \citep{Guzy:2020}. 
	
	\smallskip
	\noindent\textbf{Learning from data streams}. There are several different learning modes in which a stream can be handled \citep{Gama:2014}. A classifier may process the instances in an incremental way using mini-batches or in an online way, utilizing them one-by-one and only once \citep{Ramirez-Gallego:2018}. It may use a sliding window to track the most current concepts or a decay factor for smooth forgetting. The streaming classifiers can also be categorized into blind methods, which assume that data is dynamic and which keep updating a model all the time, or into informed ones, which react to changes based on indications of a concept drift detector \citep{Barros:2018}. In the latter case, one has to keep in mind that changes may have different characteristics, for example, they may be abrupt, gradual or recurring, therefore, different approaches may more adequate for different scenarios \citep{Lu:2019}. 
	
	\smallskip
	\noindent\textbf{Shallow learning.} Most of the state-of-the-art algorithms for streaming classification learn in a shallow way without utilizing an in-depth feature engineering process \citep{Krawczyk:2017}. 
	
	The most important group of the classifiers are those based on online Hoeffding trees (HT), designed explicitly for drifting data \citep{Hulten:2001}. The decision trees can grow themselves as new instances arrive, using a theoretically grounded incremental split condition, based on the Hoeffding bound. In addition, their improved adaptive versions (HAT) are able to remove outdated nodes and replace them with alternative background ones, if a change on a node is detected \citep{Bifet:2009aht}. Finally, the error tracking process and change indications are supported by an efficient adaptive sliding window ADWIN \citep{Bifet:2007}, which dynamically adjusts its size in order to stay reactive to the most recent concepts. 
	
	The Hoeffding trees have been used in the implementation of Adaptive Random Forest (ARF) \citep{Gomes:2017}, which is an online version of the random forest algorithm \citep{Breiman:2001}. It utilizes online bagging, splits based on feature subspaces and replacements of the outdated trees. The ensemble exhibits very reliable performance in general, being one of the most important state-of-the-art classifiers in non-stationary streaming scenarios \citep{Cano:2020}.
	
	Many of the most popular algorithms for data streams are ensembles \citep{Krawczyk:2017}. Some of them implement a dynamic weighting of base learners in order to smoothly modify their importance during majority voting, to replace the underperforming models with new ones, or to perform a dynamic selection of the most competent learners \citep{Zyblewski:2021}. Among others, we can distinguish Dynamically Weighted Majority (DWM) \citep{Kolter:2003}, Accuracy Updated Ensemble (AUE) \citep{Brzezinski:2014} or Accuracy Weighted Ensemble (AWE) \citep{Wang:2003}. Other committees are streaming versions of widely used offline ensembles, such us Oza Bagging (OBG) and Oza Boosting (OBS) \citep{Oza:2005}, Smooth Boosting (SB) \citep{Chen:2012} or Leveraging Bagging (LB) \citep{Bifet:2010}, which can be distinguished as one of the most competitive streaming classifiers next to ARF \citep{Cano:2020}.
	
	Finally, some of the standard single classifiers, like stochastic gradient descent (SGD) or naive Bayes (NB) can be directly used in streaming scenarios, although they usually do not provide satisfying results off the shelf.
	
	\smallskip
	\noindent\textbf{Deep learning.} While deep neural networks are one of the algorithms that can used as online learners, they struggle with several problems in such settings, including catastrophic forgetting \citep{Parisi:2019} or huge numbers of configurations and hyperparameters to tune, which may be particularly challenging in dynamic environments \citep{Sahoo:2018}. Most of the works related to online deep learning focus on continual class-incremental learning \citep{Kemker:2018} or adjusting the network's structure for stationary data \citep{Li:2019}. Only very few take into consideration a possible concept drift \citep{Sahoo:2018}. 
	
	Still, most of the online neural networks do not apply any dedicated adaptation of the convolutional layers. Some generic exceptions can be found mainly for applications \citep{Zhou:2017}. Also, to the best of our knowledge, adaptive recurrent networks are practically non-existent. Therefore, even though the standard CNN or RNN can be used as an online classifier, the research of true contextual deep learning from non-stationary data streams is currently very limited.
	
	Finally, it may be surprising that despite of the tremendous success of streaming decision trees, they have never been utilized as a part of an online deep learning paradigm, which, nota bene, does not have to solely identified with neural networks. In fact, a deep forest algorithm (gcForest) has been already proposed for offline settings \citep{Zhou:2017df}, presenting an alternative approach to tackling the problem of learning from high-dimensional contextual data. In this work, we attempt to fill the obvious gap.
	
	\section{Adaptive Deep Forest}
	\label{sec:adf}
	
	\subsection{Motivation}
	The original deep forest exhibits capabilities of learning from contextual data, like texts or images, by employing some kind of a hierarchical learning from different views, based on multi-grained scanning and cascades of random forests \citep{Zhou:2017df}. The algorithm has been successfully applied to different domains, proving its general usability \citep{Shen:2018}. 
	
	However, deep forest is a solely batch-based offline algorithm, unsuitable for more general problem of incremental learning and adaptation to dynamic concepts. Nevertheless, the following points explain why it can and should be changed.
	
	\begin{itemize}
		\smallskip
		\item[--] The main obstacle lies in base learners, which are random forests consisting of standard offline decision trees \citep{Breiman:2001}. This actually very fortunate. As we mentioned in the previous section, there are very reliable and well-established tree-based classifiers dedicated to streaming data -- HAT and ARF \citep{Cano:2020}. While the algorithms have been proven to be one of the best classifiers for non-stationary data, they have been designed as shallow learners, without explicit procedures for handling more complex problems. Therefore, it seems completely natural to think about replacing the standard random forests in gcForest with ARF to combine the strengths of both algorithms and, as a result, enable deep online adaptive learning from drifting contextual data. 
		
		\smallskip
		\item[--] Since Hoeffding Trees have been proven to be asymptotically close with offline counterparts \citep{Hulten:2001}, there is a chance that by using them we should be able to maintain the effectiveness of deep forest, while making it capable of online adaptation. 
		
		\smallskip
		\item[--] Finally, as the mentioned tree-based models have been exclusively designed for streaming data, there is also a chance that such a deep model may be competitive against deep neural networks, which struggle with several problems while learning in an online manner \citep{Parisi:2019}, and offer a different set of capabilities, like one-pass learning, less or lack of hyper-parameters and natural interpretability.
	\end{itemize}
	
	\subsection{Algorithm}
	Motivated by the presented observations and assumptions, we propose the Adaptive Deep Forest (ADF) algorithm -- an online version of gcForest for handling evolving high-dimensional contextual streaming data.
	
	\smallskip
	\noindent\textbf{Model}. The ADF can be described as a model consisting of the input layer ${\mathbf{I} = \{I_1, I_2, ..., I_d\}}$ and the cascade layer ${\mathbf{C} = \{C_1, C_2, ..., C_d\}}$, where $I_i$ and $C_i$ are their sublayers and $d$ defines the depth of the model. Each of the sublayers is a set of ARF classifiers. We define them as ${I_i = \{\Psi_i^{(1)}, \Psi_i^{(2)}, ..., \Psi_i^{(n)}\}}$ and $C_i = \{\Phi_i^{(1)}, \Phi_i^{(2)}, ..., \Phi_i^{(m)}\}$, where $n, m \in \mathbb{N}$ determine how many forests are in a sublayer. 
	
	For the cascade modules we also define weights ${\alpha=\{\alpha_1, \alpha_2, ..., \alpha_m\}}$, which measure the quality of all sublayers, and $\beta_i=\{\beta_i^{(1)}, \beta_i^{(2)}, ..., \beta_i^{(t)}\}$, which indicate the quality of forests in a sublayer $C_i$. In this work, we decided to use reliable ADWIN-based accuracy \citep{Bifet:2007}. By doing so, we ensure that we always have an up-to-date reactive measure and get rid of an additional parameter like a sliding window size or decay factor. 
	
	\begin{figure}[htb]
		\centering
		\includegraphics[width=.99\linewidth]{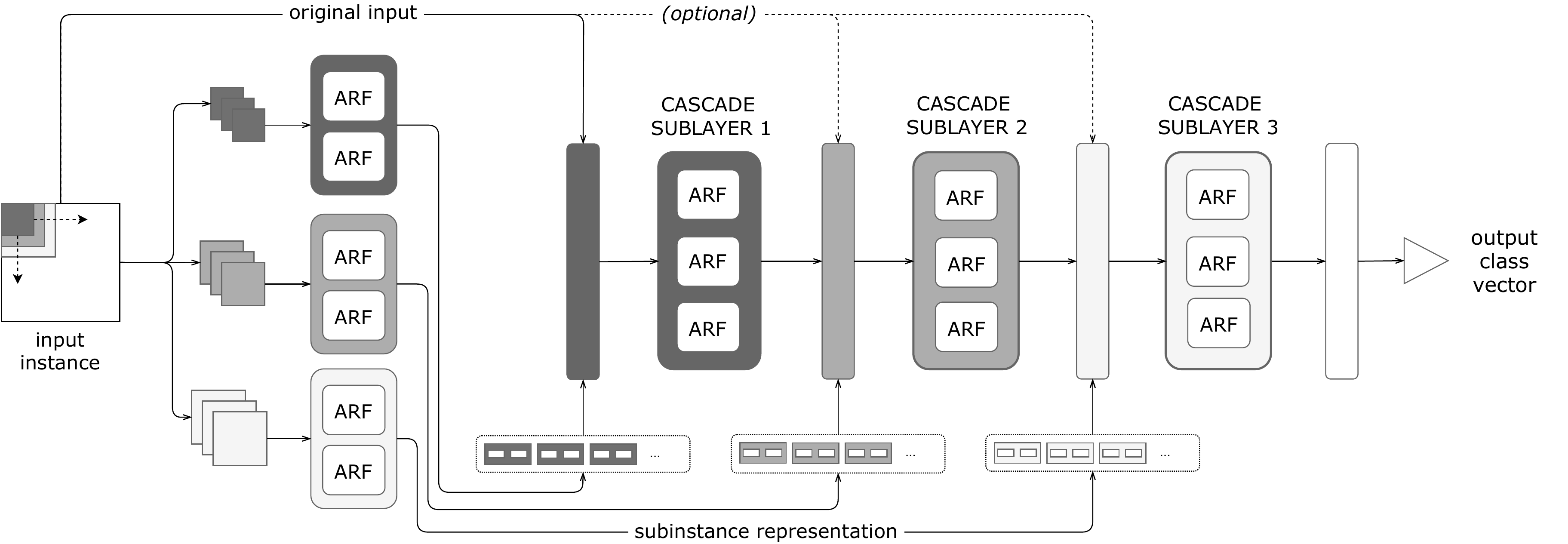}
		\caption{The illustration of ADF.}
		\label{fig:adf}
	\end{figure}
	
	\smallskip
	\noindent\textbf{Input layer}. The procedure of acquiring a class vector for a new instance is described in Alg. \ref{alg:input}-\ref{alg:output} and illustrated in Fig.\ref{fig:adf}. In the first step (Alg. \ref{alg:input}), an incoming instance $X$ is transformed by the input layer $\mathbf{I}$ into multi-grained representations $R$. At the beginning, we run the scanning operation, which extracts subinstances $X_i$ for each depth $i$. The $scan(X,d)$ function uses a sliding window for this purpose. Its size for vectors is equal to $s_{vec}=|X| / 2^{d-i+1}$, where $|X|$ is a number of attributes. For square matrices we define a square window which side has size $s_{mat}=\sqrt{|X|} / 2^{d-i+1}$. We do not consider rectangle cases. One can easily notice that preceding sublayers will generate more and smaller (fine-grained) instances than subsequent ones. Due to significant runtime limitations (streaming scenarios) we use only disconnected slides, which means that stride is always equal to the size of the window. As a result, if the input instance has a size based on the power of 2, we will get $2^{d-i+1}$ subinstances from a vector and $4^{d-i+1}$ from a matrix for depth $i \in \langle1,d\rangle$, which can be generalized to $2^{D(d-i+1)}$, where $D$ is dimensionality of an input. For different sizes we apply a simple padding with zeros.
	
	\begin{algorithm}[h]
		\BlankLine
		\KwData{incoming instance $X$}
		\KwInit{$R \gets \{\}$}
		\KwResult{depth representations $R$}
		\BlankLine
		
		extract instances for each depth: $\{X_1, X_2, ..., X_d\} \gets scan(X,d)$\;
		\BlankLine
		
		\For{$i\gets1$ \KwTo $d$}{
			$R_i \gets \{\}$\;
			\For{$x_{sub} \in X_i$}{
				$r \gets \{\Psi^{(1)}_i(x_{sub}), \Psi^{(2)}_i(x_{sub}), ..., \Psi^{(n)}_i(x_{sub})\}$\;
				$R_i \gets R_i \cup r$\;
			}
			$R \gets R \cup R_i$\;
		}
		
		\Return $R$	
		
		\caption{The input layer algorithm.}
		\label{alg:input}
	\end{algorithm}
	
	After the extraction of the subinstances from the original input, they are transformed using corresponding input sublayers. Subinstances $x_{sub} \in X_i$ for depth $i$ are processed by a sublayer $I_i$, therefore we can say that different sublayers may specialize in specific perspectives. Each subinstance $x_{sub}$ is turned into a partial representation $r$ by ARF classifiers belonging to the sublayer. They return $n$ class vectors for each $x_{sub}$, which are then added to $R_i$, forming the final depth representation. At the end of the input processing, we have $d$ representations of the original instance in $R$. 
	
	The size of a representation for a depth $i$ is equal to $|R_i| = |X_i| \times c \times n$, where $|X_i|$ is approximately equal to $2^{D(d-i+1)}$ and $c$ is a number of classes. For example, given a 32x32 image with $c=10$, $d=3$ and $n=20$, we will get $R_1 = 64 \times 10 \times 20 = 12800$, $R_2 = 3200$ and $R_3 = 800$. It is very important to note here that if we used a sliding window with a stride equal to 1, for $i=1$ we would get $29\times29=841$ subinstances and as a consequence, the size of the representation would increase to $|R_i|=168200$. That would cause a significant slowdown due to much more transformations in the current layer and heavy computations for extremely highly dimensional data in the cascade layer. Since in this work we focus only on a sequential implementation, we have to keep these numbers under strict control.
	
	\smallskip
	\noindent\textbf{Cascade layer}. In the next step, the initial input $X$ and acquired representations $R$ are forwarded to the cascade layer (Alg. \ref{alg:cascade}). The main purpose of this module is to obtain the most reliable extended class vector $y^{(best)}$, based on the deep information about the original instance. The sequential processing in this layer is rather straightforward. Analogously to the input layer, each representation $R_i$ is processed by a corresponding sublayer $C_i$ for a given depth $i$. For each sublayer $C_i$, an extended input $x_{in}$ is prepared, using three sources of information: output from the previous sublayer $y$, depth representation $R_i$ and original input $X$. The last one is optional for inner sublayers and we actually found out that omitting it may improve performance for some streams. The input $x_{in}$ is transformed into $m$ class vectors by the ARF classifiers in the sublayer, forming the sublayer output $y$.
	
	The size of an intermediate input for a sublayer $C_i$, where $i > 1$, is equal to: $|x_{in}| = |y| + |R_i| + |X|$, where $y=c \times m$ and $|X|$ is optional for inner sublayers. For example, given a 1024-word text with $c=10$, $d=5$ and having $m=n=20$ adaptive forests in a sublayer, we will get $|x_{in}| = 1024 + 6400 + 10 \times 20 = 7624$, if we include the original instance.
	
	\begin{algorithm}[h]
		\BlankLine
		\KwData{original instance $X$, depth representations $R$}
		\KwInit{$y \gets \{\}$, $y^{(best)} \gets \{\}$, $\alpha^{(best)} \gets -\infty$}
		\KwResult{best output $y^{(best)}$}
		\BlankLine
		
		\For{$i\gets1$ \KwTo $d$}{
			$x_{in} \gets y \cup R_i \cup X$\;
			$y \gets \{\Phi^{(1)}_i(x_{in}), \Phi^{(2)}_i(x_{in}), ..., \Phi^{(m)}_i(x_{in})\}$\;
			\BlankLine
			\uIf{$\alpha_i > \alpha^{(best)}$}{
				$y^{(best)} \gets y$\;
				$\alpha^{(best)} \gets \alpha_i$\;
			}
		}
		
		\Return $y^{(best)}$	
		
		\caption{The cascade layer algorithm.}
		\label{alg:cascade}
	\end{algorithm}
	
	It is supposed that in this layer, by the hierarchical processing of different grains, the algorithm may be able to capture some structured, contextual information about the original input. However, since we have to deal with an online scenario, we are not able to choose a fixed architecture using a specified depth. In fact, one cannot say for sure that the last sublayer will always provide the most reliable output. The optimal depth may be different not only for different streams, but also for different parts of the same stream (concept drift, incremental learning). Therefore, since each sublayer returns a vector that can be aggregated into a class vector, we implemented a dynamic depth selection, which picks the output of the most effective cascade sublayer $y^{(best)}$ as the final one. The decision is made based on weights $\alpha_i$ of the sublayers.
	
	\begin{algorithm}[ht]
		\BlankLine
		\KwData{best subplayer output $y^{(best)}$}
		\KwInit{$\hat{y} \gets \{\}$}
		\KwResult{prediction vector $\hat{y}$}
		\BlankLine
		
		\For{$k\gets1$ \KwTo $C$}{
			$\hat{y} \gets \hat{y} \cup \dfrac{\sum_{j=1}^{m}\beta_j^{(best)}y_{(j-1)C}^{(best)}}{\sum_{j=1}^{m}\beta^{(best)}_j}$\;
		}
		
		\Return $\hat{y}$	
		
		\caption{The output aggregation algorithm.}
		\label{alg:output}
	\end{algorithm}
	
	\smallskip
	\noindent\textbf{Final output}. Given the most reliable output $y^{(best)}$ we aggregate it in order to obtain the final class vector (Alg. \ref{alg:output}). Because the obtained output is a result of concatenated predictions made by multiple adaptive forests, we can see the final aggregation as an ensemble voting. Since we track the quality of all forest $\beta_i$ for a given sublayer $C_i$, we can apply a weighted voting, which should further improve the overall performance.
	
	Finally, all the operations are complemented with updating the ARFs models and weights during the training phase.
	
	\section{Experimental study}
	\label{sec:exp}
	
In this section, we present experiments conducted to empirically analyze the proposed algorithm. The experimental study was designed to answer the following research questions:

\begin{itemize}

\item \textbf{RQ1:} What are the contributions of individual sublayers of Adaptive Deep Forest?

\smallskip
\item \textbf{RQ2:} How does the size of ensemble in each sublayer and the number of used forests affect the predictive performance of Adaptive Deep Forest?

\smallskip
\item \textbf{RQ3:} Is Adaptive Deep Forest capable of outperforming state-of-the-art online classifiers, especially for complex and high-dimensional data streams?

\end{itemize}

This allows us to divide experiments and the discussion of results into two parts. Firstly, we provide some insights into the internal learning process of ADF, showing what we can expect from different configurations. Secondly, we evaluate the predictive performance of ADF in the context of different data streams and in comparison with state-of-the-art classifiers dedicated to streaming scenarios. All the experiments along with the source code of the algorithm can be found in our repository\footnote{\label{website1}\href{https://github.com/adf-rep/adf-20}{github.com/adf-rep/adf-20}}, which we have made publicly available in order to improve the reproducibility of our work.
	
	\subsection{Data}
	
	In our experiments, we used two types of data -- 9 standard data streams (SHALLOW), commonly used while benchmarking shallow adaptive algorithms, and 12 popular contextual streams, including 8 visual ones (DEEP-2D), 3 textual and 1 time series (DEEP-1D). All of the 21 data streams are described in Tab. \ref{tab:data}. More information can be found in the repository.
	
	\begin{table}[htb]
		\caption{Summary of the used data streams.}
		\centering
		\scalebox{0.9}{
			\begin{tabular}[H]{c|lccc}
				\toprule
				\textbf{Type} & \textbf{Name} & \textbf{\#Instances} & \textbf{\#Attributes} & \textbf{\#Classes}\\
				\midrule
				\parbox[t]{2mm}{\multirow{9}{*}{\rotatebox[origin=c]{90}{SHALLOW}}} & ACTIVITY & 10 853 & 43 & 8\\
				& CONNECT4 & 67 557 & 42 & 3\\
				& COVER & 581 012 & 54 & 7\\
				& EEG & 14 980 & 14 & 2\\
				& ELEC & 45 312 & 8 & 2\\
				& GAS & 13 910 & 128 & 6\\
				& POKER & 829 201 & 10 & 10\\	 		
				& SPAM & 9 324 & 499 & 2\\
				& WEATHER & 18 158 & 8 & 2\\
				\midrule
				\parbox[t]{2mm}{\multirow{4}{*}{\rotatebox[origin=c]{90}{DEEP-1D}}} & AGNEWS & 60 000 & 169 & 4\\
				& BBC & 66 750 & 1 000 & 5\\
				& SOGOU & 60 000 & 1 500 & 5\\
				& SEMG & 54 000 & 3 000 & 6\\
				\midrule
				\parbox[t]{2mm}{\multirow{8}{*}{\rotatebox[origin=c]{90}{DEEP-2D}}} & CIFAR10 & 120 000 & 1024 (32x32) & 10\\
				& CMATER & 40 000 & 1024 (32x32) & 10\\
				& DOGSvCATS & 50 000 & 1024 (32x32) & 2\\	 		
				& FASHION & 140 000 & 784 (28x28) & 10\\
				& IMGNETTE & 37 876 & 4 096 (64x64) & 10\\
				& INTEL & 11 2272 & 1024 (32x32) & 6\\
				& MALARIA & 55 116 & 1024 (32x32) & 2\\
				& MNIST & 140 000 & 784 (28x28) & 10\\
				\bottomrule
		\end{tabular}}
		\label{tab:data}
		\vspace{0.3cm}
	\end{table}
	
	Since the proposed classifier is designed for non-stationary data, we ensured that all of the streams are dynamic. While the shallow ones consist of drifts themselves, the original deep sets are stationary, therefore, we had to extend them with a semi-synthetic concept drift. We did simply by doubling their size, shuffling and shifting classes in the middle of each stream. For the synthesis of a change, we utilized the commonly used sigmoidal formula \citep{Bifet:2018} from MOA, setting the width of a drift equal to 10\% of a whole stream (after extending it), simulating a moderate incremental drift.
	
	\subsection{Set-up}
	
	Let us introduce the most important details of the settings used in the experiments. For more specific details please refer to the introduced repository.
	
	\smallskip
	\noindent\textbf{Overview}. In the first part, we investigated the quality of the cascade sublayers, checking how the predictive performance changes with depth. We also evaluated how increasing the number of trees and forests affects the quality of classification and runtime. Then, we conducted the main evaluation by running our and all other algorithms on the given benchmarks.
	
	\smallskip
	\noindent\textbf{Metrics}. While measuring the predictive performance, we tracked the prequential accuracy and kappa value within ADWIN, as recommended in \citep{Bifet2015:eval}, to capture temporal quality as a series. We also reported the average values for whole streams \citep{Bifet:2018}. For the series presentation, we included only accuracy, since kappa within a sliding window tends to be much noisier. While measuring runtime, we tracked the prediction and update time, reporting all of them or only the total time, when it was justified. Finally, for the statistical analysis of the final comparison, we used the Bonferroni-Dunn ranking test with significance level $\alpha_t = 0.05$.
	
	\smallskip
	\noindent\textbf{Classifiers}. As referential classifiers we used all the state-of-the-art streaming algorithms presented in Sec. \ref{sec:stream}, commonly used in this domain: HAT, SGD, NB, LB, OBG, OBS, SB, DWM, AUC, AWE, LNSE and ARF, treating the last one as the main baseline, upon which the whole ADF has been built. We also distinguished the cascade layer without the input layer and depth representations as a separate classifier (CARF) in order to analyze the influence of the contextual multi-grained scanning in the given scenarios.
	
	\smallskip
	\noindent\textbf{Parameters}. While evaluating different numbers of trees and forests, we checked $T=\{10,20,30,40,50\}$ in each sublayer for the former, and $F=\{1,2,3,4,5\}$ for the latter. We set fixed $F=2$ in the first case and $T=20$ in the second one. Larger values were not feasible due to the sequential implementation. Because of the same reason, in this part of experiments, we used only the first half of each deep stream. For visual streams we considered the maximum reasonable, with respect to the size of images, depth $d=4$ and for the 1D ones we set $d=5$. For the shallow streams with more than 32 attributes, we used $d=4$ and for the rest, which had less than 16 features, $d=2$. Based on the observations from the initial phase, during the final comparison we chose $T=25$ and $F=2$ for 2D data and $T=40$ for 1D streams, keeping $d$ as it was. We were not appending the original instance for inner cascade sublayers, as we did not observe any significant improvement, except for CIFAR10 and IMGNETTE. The CARF algorithm used analogous numbers, but in this case we activated appending, since there was no depth representation vectors passed to sublayers. ARF used the same numbers of trees. Finally, all the remaining classifiers used default settings defined in MOA.
	
	\subsection{Results and discussion}
	
\begin{figure}[htb]
		\centering
		\begin{subfigure}{0.32\textwidth}
			\centering
			\includegraphics[width=\linewidth]{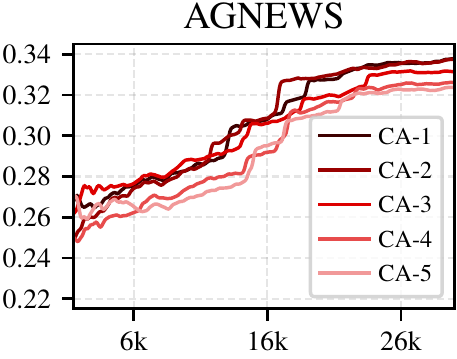}
		\end{subfigure}
		\begin{subfigure}{0.32\textwidth}
			\centering
			\includegraphics[width=\linewidth]{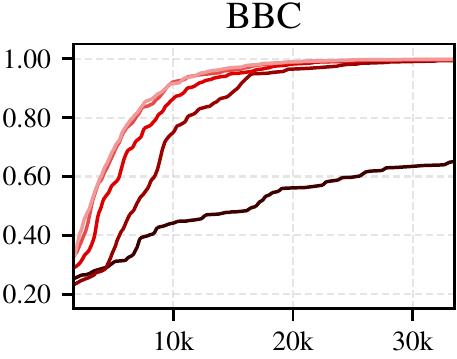}
		\end{subfigure}
		\begin{subfigure}{0.32\textwidth}
			\centering
			\includegraphics[width=\linewidth]{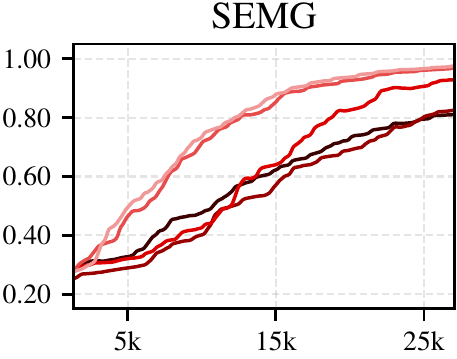}
		\end{subfigure}\\
		\vspace{3pt}
		\begin{subfigure}{0.32\textwidth}
			\centering
			\includegraphics[width=\linewidth]{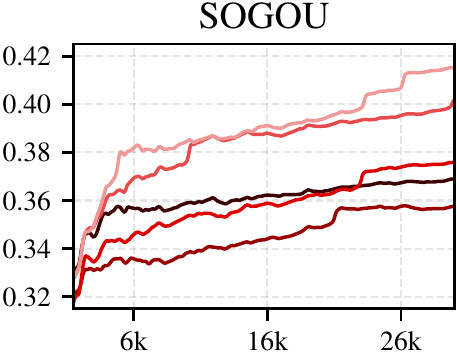}
		\end{subfigure}
		\begin{subfigure}{0.32\textwidth}
			\centering
			\includegraphics[width=\linewidth]{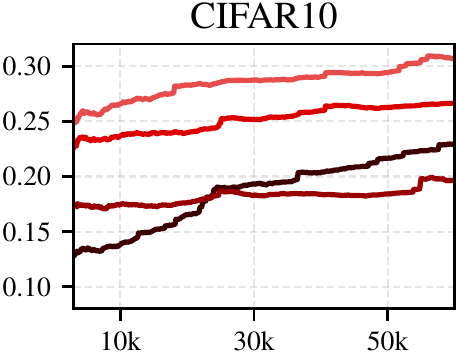}
		\end{subfigure}
		\begin{subfigure}{0.32\textwidth}
			\centering
			\includegraphics[width=\linewidth]{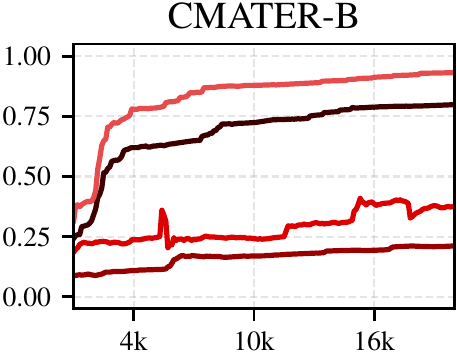}
		\end{subfigure}\\
		\vspace{3pt}
		\begin{subfigure}{0.32\textwidth}
			\centering
			\includegraphics[width=\linewidth]{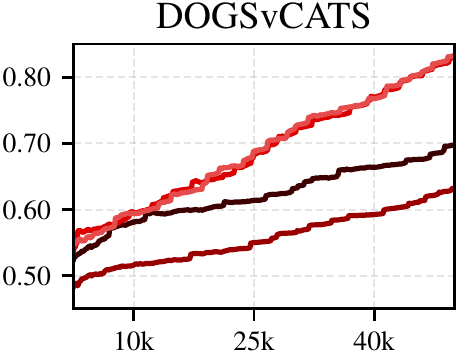}
		\end{subfigure}
		\begin{subfigure}{0.32\textwidth}
			\centering
			\includegraphics[width=\linewidth]{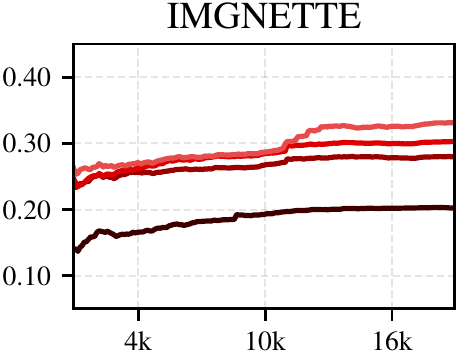}
		\end{subfigure}
		\begin{subfigure}{0.32\textwidth}
			\centering
			\includegraphics[width=\linewidth]{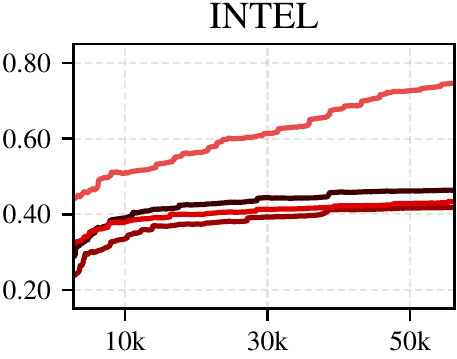}
		\end{subfigure}\\
		\vspace{3pt}
		\begin{subfigure}{0.32\textwidth}
			\centering
			\includegraphics[width=\linewidth]{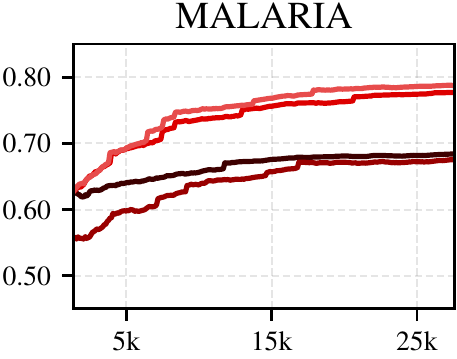}
		\end{subfigure}
		\begin{subfigure}{0.32\textwidth}
			\centering
			\includegraphics[width=\linewidth]{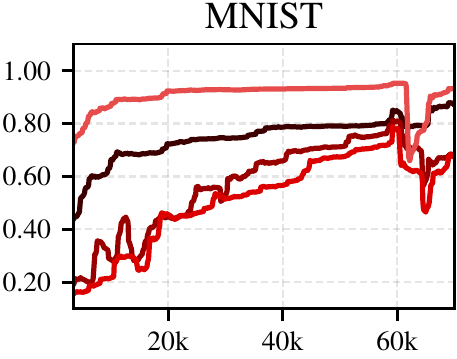}
		\end{subfigure}
		\begin{subfigure}{0.32\textwidth}
			\centering
			\includegraphics[width=\linewidth]{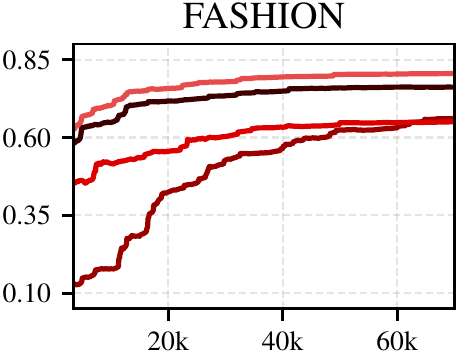}
		\end{subfigure}
		\caption{Accuracy (weights) per cascade sublayer for the used concepts.}
		\label{fig:weights}
		\vspace{0.2cm}
	\end{figure}

	\noindent\textbf{Deep sublayers.} As the first part, let us analyze how the predictive performance changes (weights $\alpha$) while going through sublayers in the deep cascade layer. Fig. \ref{fig:weights} illustrates the values of the weights for some of the 1D and 2D concepts used in the experiments. We can clearly see that in almost all cases the last sublayer turned out to produce the most accurate prediction, providing from about 0.7 in accuracy gain for CMATER-B, compared with the worst sublayer, to slightly less than 0.2 for such streams like SEMG or FASHION. The only exception can be seen for AGNEWS. Most interestingly, only for BBC and SEMG we can notice that the quality of prediction was increasing hierarchically, from the first sublayer (CA-1) to the last one (CA-5). In all other cases, this cannot be claimed. What we can see instead is that very often CA-1 was better than CA-2 and CA-3. However, at the same time, even if the latter sublayers performed worse than the earlier one, ultimately, the last sublayer (CA-5 or CA-4) was the most effective one (\textbf{RQ1 answered}). 
	
\begin{figure}[htb]
		\centering
		\begin{subfigure}{0.38\columnwidth}
			\centering
			\includegraphics[width=\columnwidth]{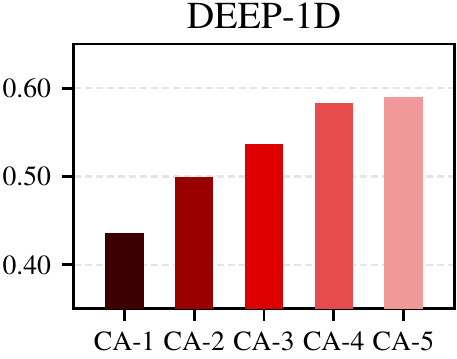}
		\end{subfigure}
		\hspace*{10pt}
		\begin{subfigure}{0.38\columnwidth}
			\centering
			\includegraphics[width=\columnwidth]{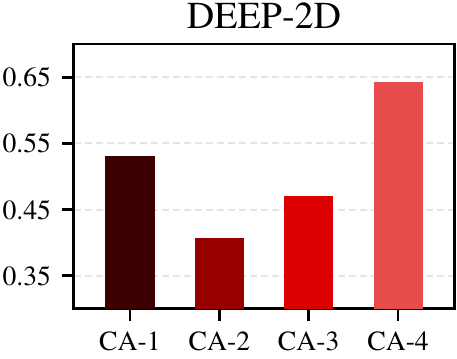}
		\end{subfigure}
		\caption{Average accuracy (weight) per cascade sublayer.}
		\label{fig:weights_avg}
		\vspace{0.2cm}
	\end{figure}

	It is possible that while the internal sublayers are not able to produce good predictions on their own, they contribute positively with their intermediate output representations to the final decision, allowing for some form of deep learning. A more in-depth investigation is required to confirm this hypothesis. Furthermore, while there is no strict hierarchy of the sublayers, the dynamic selection of the best output, introduced for the cascade layer, may turn out to be useful. On the other hand, it is worth noting that the hierarchical characteristics were observed for the streams from which ADF learned very effectively (BBC, SEMG) -- it suggests that the better ADF captures class concepts, the more practical cascading may be. On average (Fig. \ref{fig:weights_avg}) it can be observed for 1D streams, but not for the 2D ones.
	
	Finally, as one can notice in most of the cases, ADF was able to incrementally learn given concepts, especially for the best sublayers, converging smoothly for most of the cases. Learning for AGNEWS or CIFAR10 was distinctively slower than for the other streams. Nevertheless, the results generally indicate the validity of the online implementation of deep forest.
	
	\smallskip
	\noindent\textbf{Trees and forests.} When looking at the results for different numbers of trees (Tab. \ref{tab:num trees}) and forests (Tab. \ref{tab:num forests}) for each ARF and sublayer, we can easily notice that adding more models to the whole structure resulted in notable performance improvements only between 10 and 20 trees or 1 and 2 forests, with significant changes only for 1D streams. At the other hand, increasing the size of sublayers led to a proportional increase in runtime, therefore, the obtained trade-off is extremely ineffective.
	
	\begin{table}[htb]
		\caption{Results for different numbers of trees.}
		\centering
		\begin{subtable}{\columnwidth}
			\centering	
			\subcaption{Prequential accuracy}
			\scalebox{0.95}{\begin{tabular}[]{>{\arraybackslash} m{1.5cm} >{\centering\arraybackslash} m{1cm} >{\centering\arraybackslash} m{1cm} >{\centering\arraybackslash} m{1cm} >{\centering\arraybackslash} m{1cm} >{\centering\arraybackslash} m{1cm}}
				\toprule	
				\textbf{Streams} & \textbf{10} & \textbf{20} & \textbf{30} & \textbf{40} & \textbf{50}\\
				\midrule
				SHALLOW&0.8981&0.9017&0.9028&0.9034&0.9030\\
				DEEP-1D&0.5156&0.5615&0.5757&0.5785&0.5813\\
				DEEP-2D&0.6177&0.6312&0.6375&0.6374&0.6401\\
				\midrule
				Average&0.6771&0.6981&0.7053&0.7064&0.7081\\
				\bottomrule
			\end{tabular}}
		\end{subtable}\\
		\vspace*{0.25cm}
		\begin{subtable}{\columnwidth}
			\centering	
			\subcaption{Kappa}
			\scalebox{0.95}{\begin{tabular}[]{>{\arraybackslash} m{1.5cm} >{\centering\arraybackslash} m{1cm} >{\centering\arraybackslash} m{1cm} >{\centering\arraybackslash} m{1cm} >{\centering\arraybackslash} m{1cm} >{\centering\arraybackslash} m{1cm}}
				\toprule	
				\textbf{Streams} & \textbf{10} & \textbf{20} & \textbf{30} & \textbf{40} & \textbf{50}\\
				\midrule
				SHALLOW & 0.7890 & 0.7961 & 0.7990 & 0.8001 & 0.7999\\
				DEEP-1D & 0.3567 & 0.4131 & 0.4311 & 0.4343 & 0.4379\\
				DEEP-2D & 0.5689 & 0.5843 & 0.5914 & 0.5914 & 0.5944\\
				\midrule
				Average & 0.5715 & 0.5978 & 0.6072 & 0.6086 & 0.6107\\
				\bottomrule
			\end{tabular}}
		\end{subtable}\\
		\vspace*{0.25cm}
		\begin{subtable}{\columnwidth}
			\centering	
			\subcaption{Time [ms]}
			\scalebox{0.95}{\begin{tabular}[]{>{\arraybackslash} m{1.5cm} >{\centering\arraybackslash} m{1cm} >{\centering\arraybackslash} m{1cm} >{\centering\arraybackslash} m{1cm} >{\centering\arraybackslash} m{1cm} >{\centering\arraybackslash} m{1cm}}
				\toprule	
				\textbf{Streams} & \textbf{10} & \textbf{20} & \textbf{30} & \textbf{40} & \textbf{50}\\
				\midrule
				SHALLOW & 7 & 15 & 23 & 31 & 39\\
				DEEP-1D & 64 & 147 & 208 & 279 & 320\\
				DEEP-2D & 80 & 161 & 246 & 305 & 368\\
				\midrule
				Average & 50 & 107 & 159 & 205 & 242\\
				\bottomrule
				\\
			\end{tabular}}
		\end{subtable}
		\label{tab:num trees}
	\end{table}
	
	\begin{table}[htb]
		\caption{Results for different numbers of forests.}
		\centering
		\begin{subtable}{\columnwidth}
			\centering	
			\subcaption{Prequential accuracy}
			\scalebox{0.95}{\begin{tabular}[]{>{\arraybackslash} m{1.5cm} >{\centering\arraybackslash} m{1cm} >{\centering\arraybackslash} m{1cm} >{\centering\arraybackslash} m{1cm} >{\centering\arraybackslash} m{1cm} >{\centering\arraybackslash} m{1cm}}
				\toprule	
				\textbf{Streams} & \textbf{1} & \textbf{2} & \textbf{3} & \textbf{4} & \textbf{5}\\
				\midrule
				SHALLOW & 0.7997 & 0.7954 & 0.7950 & 0.7934 & 0.7916\\
				DEEP-1D & 0.3779 & 0.4131 & 0.4269 & 0.4321 & 0.4375\\
				DEEP-2D & 0.5759 & 0.5843 & 0.5853 & 0.5841 & 0.5797\\
				\midrule
				Average & 0.5845 & 0.5976 & 0.6024 & 0.6032 & 0.6030\\
				\bottomrule
			\end{tabular}}
		\end{subtable}\\
		\vspace*{0.25cm}
		\begin{subtable}{\columnwidth}
			\centering	
			\subcaption{Kappa}
			\scalebox{0.95}{\begin{tabular}[]{>{\arraybackslash} m{1.5cm} >{\centering\arraybackslash} m{1cm} >{\centering\arraybackslash} m{1cm} >{\centering\arraybackslash} m{1cm} >{\centering\arraybackslash} m{1cm} >{\centering\arraybackslash} m{1cm}}
				\toprule	
				\textbf{Streams} & \textbf{1} & \textbf{2} & \textbf{3} & \textbf{4} & \textbf{5}\\
				\midrule
				SHALLOW & 0.7997 & 0.7954 & 0.7950 & 0.7934 & 0.7916\\
				DEEP-1D & 0.3779 & 0.4131 & 0.4269 & 0.4321 & 0.4375\\
				DEEP-2D & 0.5759 & 0.5843 & 0.5853 & 0.5841 & 0.5797\\
				\midrule
				Average & 0.5845 & 0.5976 & 0.6024 & 0.6032 & 0.6030\\
				\bottomrule
			\end{tabular}}
		\end{subtable}\\
		\vspace*{0.25cm}
		\begin{subtable}{\columnwidth}
			\centering	
			\subcaption{Time [ms]}
			\scalebox{0.95}{\begin{tabular}[]{>{\arraybackslash} m{1.5cm} >{\centering\arraybackslash} m{1cm} >{\centering\arraybackslash} m{1cm} >{\centering\arraybackslash} m{1cm} >{\centering\arraybackslash} m{1cm} >{\centering\arraybackslash} m{1cm}}
				\toprule	
				\textbf{Streams} & \textbf{1} & \textbf{2} & \textbf{3} & \textbf{4} & \textbf{5}\\
				\midrule
				SHALLOW & 4 & 8 & 13 & 18 & 24\\
				DEEP-1D & 57 & 115 & 206 & 272 & 303\\
				DEEP-2D & 67 & 137 & 209 & 298 & 411\\
				\midrule
				Average & 43 & 87 & 143 & 196 & 246\\
				\bottomrule
				\\
			\end{tabular}}
		\end{subtable}
		\label{tab:num forests}
		\vspace{0.2cm}
	\end{table}
	
	A possible explanation for the observed results is that we do not provide enough diversity in intermediate data used by sublayers. Firstly, the relatively large stride for the scanning sliding window in the input layer (equal to the size of the window), may result in producing not long enough depth representations for trees in cascade sublayers. As a consequence, when increasing the number of trees, the random feature subspaces used on their nodes may turn out to be more severely correlated among the models, resulting in similar base learners. That obviously will not provide any additional gain. For a similar reason, trees in preceding input sublayers may be more correlated with each other (smaller subinstances) than those in subsequent ones.
	
	\begin{table}[htb]
		\caption{Average prequential accuracy for all classifiers and data streams.}
		\centering
		\scalebox{0.85}{\begin{tabular}[]{>{\arraybackslash} m{1.7cm} >{\centering\arraybackslash} m{0.9cm} >{\centering\arraybackslash} m{0.9cm} >{\centering\arraybackslash} m{0.9cm} >{\centering\arraybackslash} m{0.9cm} >{\centering\arraybackslash} m{0.9cm} >{\centering\arraybackslash} m{0.9cm} >{\centering\arraybackslash} m{0.9cm}}
				\toprule	
				\textbf{Stream} & \textbf{HAT} & \textbf{SGD} & \textbf{NB} & \textbf{LB} & \textbf{OBG} & \textbf{OBS} & \textbf{SB}\\
				\midrule
				ACTIVITY & 0.7105 & 0.5691 & 0.6905 & 0.8329 & 0.8112 & 0.7242 & 0.7097\\
				CONNECT4 & 0.7243 & 0.6737 & 0.6923 & \textbf{0.7774} & 0.7478 & 0.7763 & 0.7621\\
				COVER & 0.8189 & 0.6260 & 0.6052 & 0.9170 & 0.8474 & 0.8995 & 0.8398\\
				EEG & 0.8704 & \textbf{0.9985} & 0.4742 & 0.9834 & 0.9394 & 0.9694 & 0.8270\\
				ELEC & 0.8339 & 0.5758 & 0.7336 & 0.8980 & 0.8435 & 0.8673 & 0.8359\\
				GAS & 0.5866 & 0.9485 & 0.5839 & 0.8216 & 0.8058 & 0.5730 & 0.5886\\
				POKER & 0.6687 & 0.6643 & 0.5955 & 0.8760 & 0.7406 & \textbf{0.8825} & 0.8167\\
				SPAM & 0.9071 & 0.8763 & 0.9067 & 0.9609 & 0.9107 & 0.9646 & 0.9130\\
				WEATHER & 0.7353 & 0.6780 & 0.6922 & 0.7811 & 0.7501 & 0.7687 & 0.7495\\
				\midrule
				AGNEWS & 0.2574 & 0.2555 & 0.2516 & 0.2562 & 0.2559 & 0.2494 & 0.2583\\
				BBC & 0.2321 & 0.4887 & 0.2016 & 0.3092 & 0.2406 & 0.2016 & 0.2376\\
				SEMG & 0.1796 & 0.2408 & 0.2353 & 0.2219 & 0.1732 & 0.1671 & 0.1895\\
				SOGOU & 0.3089 & 0.2909 & 0.2531 & 0.3360 & 0.3094 & 0.2533 & 0.3052\\
				\midrule
				CIFAR10 & 0.2586 & 0.0999 & 0.1859 & 0.2458 & 0.2528 & 0.1003 & 0.2524\\
				CMATER & 0.6831 & 0.5158 & 0.4801 & 0.7778 & 0.7915 & 0.5268 & 0.5136\\
				DOGSvCATS & 0.5765 & 0.5367 & 0.4996 & 0.6189 & 0.5935 & 0.5888 & 0.5887\\
				FASHION & 0.3692 & 0.4723 & 0.3588 & 0.4105 & 0.3421 & 0.2699 & 0.2684\\
				IMGNETTE & 0.1792 & 0.1034 & 0.1795 & 0.1825 & 0.1864 & 0.1001 & 0.1622\\
				INTEL & 0.4232 & 0.1708 & 0.3070 & 0.4336 & 0.4306 & 0.1610 & 0.4126\\
				MALARIA & 0.5655 & 0.5148 & 0.5044 & 0.5482 & 0.5451 & 0.5205 & 0.5377\\
				MNIST & 0.6873 & 0.4771 & 0.4654 & 0.7425 & 0.7179 & 0.6878 & 0.6682\\
				\midrule
				\textbf{Average} & 0.5513 & 0.5132 & 0.4713 & 0.6158 & 0.5826 & 0.5358 & 0.5446\\
				\textbf{Rank} & 8.14 & 10.00 & 11.45 & 4.81 & 6.76 & 8.60 & 8.04\\
				\bottomrule
				\\
		\end{tabular}}
		\scalebox{0.85}{\begin{tabular}[]{>{\arraybackslash} m{1.7cm} >{\centering\arraybackslash} m{0.9cm} >{\centering\arraybackslash} m{0.9cm} >{\centering\arraybackslash} m{0.9cm} >{\centering\arraybackslash} m{0.9cm} >{\centering\arraybackslash} m{0.9cm} >{\centering\arraybackslash} m{0.9cm} >{\centering\arraybackslash} m{0.9cm}}
				\toprule	
				\textbf{Stream} & \textbf{DWM} & \textbf{AUC} & \textbf{AWE} & \textbf{LNSE} & \textbf{ARF} & \textbf{CARF} & \textbf{ADF}\\
				\midrule
				ACTIVITY & 0.8080 & 0.5973 & 0.6046 & 0.5078 & 0.8886 & 0.8961 & \textbf{0.9169}\\
				CONNECT4 & 0.7477 & 0.7062 & 0.6433 & 0.6710 & 0.7506 & 0.7593 & 0.7584\\
				COVER & 0.8290 & 0.8712 & 0.8023 & 0.6403 & 0.9260 & 0.9388 & \textbf{0.9477}\\
				EEG & 0.9188 & 0.5871 & 0.5804 & 0.5284 & 0.9866 & 0.9874 & 0.9976\\
				ELEC & 0.7966 & 0.7744 & 0.7092 & 0.7106 & 0.8885 & 0.8906 & \textbf{0.9130}\\
				GAS & 0.8540 & 0.5499 & 0.4533 & 0.4874 & 0.9590 & 0.9636 & \textbf{0.9642}\\
				POKER & 0.7291 & 0.6727 & 0.6018 & 0.5920 & 0.8203 & 0.8473 & 0.8303\\
				SPAM & 0.9089 & 0.7358 & 0.7415 & 0.7056 & 0.9664 & \textbf{0.9683} & 0.9650\\
				WEATHER & 0.7013 & 0.7534 & 0.7029 & 0.6864 & \textbf{0.7932} & 0.7903 & 0.7846\\
				\midrule
				AGNEWS & 0.2592 & 0.2542 & 0.2541 & 0.2494 & 0.2940 & 0.2968 & \textbf{0.3292}\\
				BBC & 0.2016 & 0.2695 & 0.2016 & 0.2016 & 0.6292 & 0.6848 & \textbf{0.8903}\\
				SEMG & 0.2500 & 0.1757 & 0.1671 & 0.1671 & 0.6556 & 0.7391 & \textbf{0.7771}\\
				SOGOU & 0.2531 & 0.3076 & 0.2531 & 0.2531 & 0.3824 & 0.3847 & \textbf{0.3983}\\
				\midrule
				CIFAR10 & 0.2457 & 0.2457 & 0.2465 & 0.0999 & 0.3012 & \textbf{0.3037} & 0.2891\\
				CMATER & 0.8023 & 0.7711 & 0.7188 & 0.5454 & 0.8781 & 0.8889 & \textbf{0.8896}\\
				DOGSvCATS & 0.5680 & 0.5863 & 0.5677 & 0.5608 & 0.6933 & 0.6925 & \textbf{0.7064}\\
				FASHION & 0.5922 & 0.1236 & 0.4024 & 0.4318 & 0.7492 & 0.7559 & \textbf{0.7593}\\
				IMGNETTE & 0.2226 & 0.1142 & 0.1001 & 0.1001 & \textbf{0.3171} & 0.3154 & 0.3116\\
				INTEL & 0.4315 & 0.2210 & 0.3555 & 0.3404 & 0.5900 & 0.6007 & \textbf{0.6234}\\
				MALARIA & 0.5431 & 0.5065 & 0.5007 & 0.5148 & 0.6782 & 0.6768 & \textbf{0.7543}\\
				MNIST & 0.7404 & 0.7390 & 0.7439 & 0.4485 & 0.8478 & 0.8508 & \textbf{0.8838}\\
				\midrule
				\textbf{Average} & 0.5906 & 0.5030 & 0.4929 & 0.4496 & 0.7141 & 0.7253 & \textbf{0.7471}\\
				\textbf{Rank} & 7.33 & 9.60 & 11.02 & 12.29 & 2.95 & 2.24 & \textbf{1.76}\\
				\bottomrule
				\\
		\end{tabular}}
		\label{tab:final-acc}
	\end{table}
	
	\begin{table}[htb]
		\caption{Average kappa for all classifiers and data streams.}
		\centering
		\scalebox{0.85}{\begin{tabular}[]{>{\arraybackslash} m{1.7cm} >{\centering\arraybackslash} m{0.9cm} >{\centering\arraybackslash} m{0.9cm} >{\centering\arraybackslash} m{0.9cm} >{\centering\arraybackslash} m{0.9cm} >{\centering\arraybackslash} m{0.9cm} >{\centering\arraybackslash} m{0.9cm} >{\centering\arraybackslash} m{0.9cm}}
				\toprule	
				\textbf{Stream} & \textbf{HAT} & \textbf{SGD} & \textbf{NB} & \textbf{LB} & \textbf{OBG} & \textbf{OBS} & \textbf{SB}\\
				\midrule
				ACTIVITY & 0.6304 & 0.4220 & 0.6035 & 0.7822 & 0.7542 & 0.6470 & 0.6281\\
				CONNECT4 & 0.3911 & 0.1179 & 0.2917 & 0.4996 & 0.4308 & \textbf{0.5045} & 0.4671\\
				COVER & 0.7125 & 0.3593 & 0.4091 & 0.8667 & 0.7563 & 0.8387 & 0.7413\\
				EEG & 0.7360 & 0.9969 & 0.0351 & 0.9665 & 0.8774 & 0.9382 & 0.6554\\
				ELEC & 0.6556 & 0.0016 & 0.4254 & 0.7900 & 0.6752 & 0.7277 & 0.6636\\
				GAS & 0.5090 & 0.9373 & 0.5061 & 0.7846 & 0.7657 & 0.4932 & 0.5116\\
				POKER & 0.4056 & 0.3626 & 0.2479 & 0.7753 & 0.5282 & \textbf{0.7883} & 0.6663\\
				SPAM & 0.7635 & 0.6821 & 0.7577 & 0.8972 & 0.7721 & 0.9062 & 0.7767\\
				WEATHER & 0.3741 & 0.2513 & 0.3130 & 0.4573 & 0.3505 & 0.4529 & 0.3696\\
				\midrule
				AGNEWS & 0.0088 & 0.0079 & 0.0022 & 0.0080 & 0.0072 & 0.0001 & 0.0103\\
				BBC & 0.0359 & 0.3608 & 0.0000 & 0.1320 & 0.0470 & 0.0000 & 0.0414\\
				SEMG & 0.0152 & 0.0886 & 0.0823 & 0.0661 & 0.0075 & 0.0000 & 0.0273\\
				SOGOU & 0.0409 & 0.0516 & 0.0000 & 0.0853 & 0.0284 & 0.0002 & 0.0109\\
				\midrule
				CIFAR10 & 0.1762 & 0.0000 & 0.0955 & 0.1620 & 0.1697 & 0.0005 & 0.1693\\
				CMATER & 0.6479 & 0.4620 & 0.4223 & 0.7531 & 0.7683 & 0.4743 & 0.4596\\
				DOGSvCATS & 0.1530 & 0.0734 & 0.0006 & 0.2377 & 0.1870 & 0.1777 & 0.1773\\
				FASHION & 0.2991 & 0.4137 & 0.2875 & 0.3449 & 0.2690 & 0.1887 & 0.1871\\
				IMGNETTE & 0.0875 & 0.0037 & 0.0875 & 0.0914 & 0.0958 & 0.0001 & 0.0685\\
				INTEL & 0.3072 & 0.0140 & 0.1667 & 0.3197 & 0.3160 & 0.0024 & 0.2946\\
				MALARIA & 0.1308 & 0.0297 & 0.0081 & 0.0954 & 0.0893 & 0.0400 & 0.0747\\
				MNIST & 0.6524 & 0.4169 & 0.4052 & 0.7139 & 0.6866 & 0.6531 & 0.6313\\
				\midrule
				\textbf{Average} & 0.3682 & 0.2883 & 0.2451 & 0.4680 & 0.4087 & 0.3730 & 0.3634\\
				\textbf{Rank} & 7.93 & 9.88 & 11.48 & 4.86 & 7.10 & 8.48 & 8.14\\
				\bottomrule
				\\
		\end{tabular}}
		\scalebox{0.85}{\begin{tabular}[]{>{\arraybackslash} m{1.7cm} >{\centering\arraybackslash} m{0.9cm} >{\centering\arraybackslash} m{0.9cm} >{\centering\arraybackslash} m{0.9cm} >{\centering\arraybackslash} m{0.9cm} >{\centering\arraybackslash} m{0.9cm} >{\centering\arraybackslash} m{0.9cm} >{\centering\arraybackslash} m{0.9cm}}
				\toprule	
				\textbf{Stream} & \textbf{DWM} & \textbf{AUC} & \textbf{AWE} & \textbf{LNSE} & \textbf{ARF} & \textbf{CARF} & \textbf{ADF}\\
				\midrule
				ACTIVITY & 0.7494 & 0.4762 & 0.4709 & 0.3376 & 0.8543 & 0.8653 & \textbf{0.8927}\\
				CONNECT4 & 0.4550 & 0.3332 & 0.2968 & 0.3137 & 0.4084 & 0.4590 & 0.4490\\
				COVER & 0.7286 & 0.7934 & 0.6772 & 0.4064 & 0.8799 & 0.9016 & \textbf{0.9160}\\
				EEG & 0.8353 & 0.1344 & 0.1267 & 0.0181 & 0.9729 & 0.9745 & \textbf{0.9951}\\
				ELEC & 0.5764 & 0.5325 & 0.4046 & 0.4143 & 0.7705 & 0.7756 & \textbf{0.8220}\\
				GAS & 0.8235 & 0.4586 & 0.3365 & 0.3772 & 0.9503 & 0.9558 & \textbf{0.9566}\\
				POKER & 0.5179 & 0.4045 & 0.2958 & 0.2814 & 0.6668 & 0.7227 & 0.6890\\
				SPAM & 0.7576 & 0.4414 & 0.4334 & 0.3489 & 0.9106 & \textbf{0.9166} & 0.9078\\
				WEATHER & 0.3710 & 0.3855 & 0.3563 & 0.3290 & 0.4736 & \textbf{0.4787} & 0.4699\\
				\midrule
				AGNEWS & 0.0121 & 0.0052 & 0.0052 & 0.0000 & 0.0582 & 0.0623 & \textbf{0.1056}\\
				BBC & 0.0000 & 0.0820 & 0.0000 & 0.0000 & 0.4982 & 0.5293 & \textbf{0.8628}\\
				SEMG & 0.0998 & 0.0104 & 0.0000 & 0.0000 & 0.5868 & 0.6589 & \textbf{0.7325}\\
				SOGOU & 0.0000 & 0.0124 & 0.0000 & 0.0000 & 0.1586 & 0.1627 & \textbf{0.1823}\\
				\midrule
				CIFAR10 & 0.1619 & 0.1619 & 0.1628 & 0.0000 & 0.2236 & \textbf{0.2264} & 0.2101\\
				CMATER & 0.7803 & 0.7456 & 0.6875 & 0.4949 & 0.8646 & 0.8765 & \textbf{0.8773}\\
				DOGSvCATS & 0.1359 & 0.1726 & 0.1354 & 0.1215 & 0.3865 & 0.3850 & \textbf{0.4127}\\
				FASHION & 0.5469 & 0.0262 & 0.3360 & 0.3687 & 0.7214 & 0.7288 & \textbf{0.7325}\\
				IMGNETTE & 0.1358 & 0.0156 & 0.0000 & 0.0000 & \textbf{0.2407} & 0.2387 & 0.2345\\
				INTEL & 0.3169 & 0.0715 & 0.2281 & 0.2102 & 0.5078 & 0.5207 & \textbf{0.5478}\\
				MALARIA & 0.0854 & 0.0116 & 0.0000 & 0.0285 & 0.3565 & 0.3536 & \textbf{0.5087}\\
				MNIST & 0.7114 & 0.7100 & 0.7154 & 0.3860 & 0.8308 & 0.8342 & \textbf{0.8709}\\
				\midrule
				\textbf{Average} & 0.4191 & 0.2850 & 0.2699 & 0.2113 & 0.5867 & 0.6013 & \textbf{0.6369}\\
				\textbf{Rank} & 7.14 & 9.62 & 11.02 & 12.26 & 3.10 & 2.19 & \textbf{1.81}\\
				\bottomrule
				\\
		\end{tabular}}
		\label{tab:final-kappa}
	\end{table}
	
	Secondly, all ARFs in an input sublayer use the same subinstances, therefore, even though online bagging is a part of the forests, they may still be similar to each other. This problem may be solved by using separate parts of the input data by different ARFs in the input sublayers. Nota bene, it would also speed up computations. Finally, due to the same reason, ARFs in the cascade layer may be highly correlated, since all of them take as an input the same intermediate vector, but this time it is only one input, so the scenario is probably even worse regarding compensation that bagging may provide (less combinations are possible).
	
	We suppose that using a smaller stride combined with locally separate subsampling in the input and cascade sublayers may effectively decorrelate trees and forests, enabling gains from increasing their numbers and allowing for more in-depth contextual learning with more specialized units -- that could improve the overall performance of ADF  (\textbf{RQ2 answered}). However, as mentioned in Sec. \ref{sec:adf}, decreasing stride is critical for runtime, which would suffer even more with more trees and forests as shown in Tab. \ref{tab:num trees} and Tab. \ref{tab:num forests}. We believe that an efficient parallel implementation of the algorithm is critical. One should also notice, that while the modifications add more computations, local subsampling would provide the opposite, which makes the potential improvements more feasible.
	
	\smallskip
	\noindent\textbf{Predictive performance.} In the last part of the experiments, we evaluated the proposed ADF algorithm in the context of other streaming classifiers. Tables \ref{tab:final-acc} and \ref{tab:final-kappa} present the average prequential accuracy and kappa values for all algorithms and data streams. What stands out it that ADF was the best classifier (accuracy: 0.7471; and kappa: 0.6369) in a predominant number of cases, including 5 out of 9 shallow streams, all 1D ones and 6 out of 8 visual benchmarks. The only competitive algorithms were: ARF (0.7141 and 0.5867) and CARF (0.7253 and 0.6013), upon which the whole algorithm has been designed, and LB, but mainly for shallow streams --  it rather failed at contextual data.
	
	It is remarkable that ADF was able to outperform almost all state-of-the-art streaming classifiers by at least 0.2 in kappa on average. The differences are even more significant for deep streams, for which most of the considered models utterly failed. It is also worth noting that while most of the algorithms use Hoeffding trees as default base learners, only the ARF-based avoided a severe drop in performance for textual, visual and time series data. This shows that ARF can be a much more reliable algorithm for complex high-dimensional data.
	
\begin{figure}[htb]
		\centering
		\begin{subfigure}{0.32\linewidth}
			\centering
			\includegraphics[width=\linewidth]{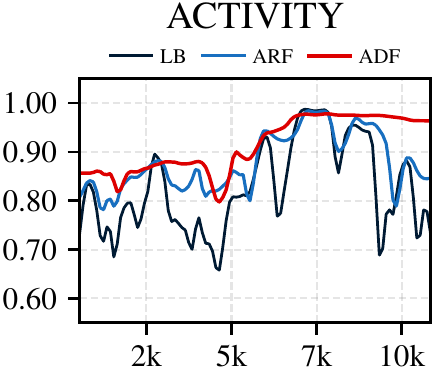}
		\end{subfigure}
		\begin{subfigure}{0.32\linewidth}
			\centering
			\includegraphics[width=\linewidth]{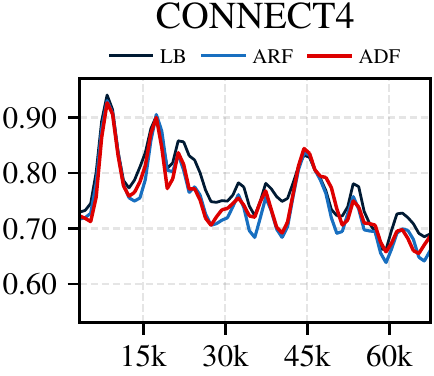}
		\end{subfigure}
		\begin{subfigure}{0.32\linewidth}
			\centering
			\includegraphics[width=\linewidth]{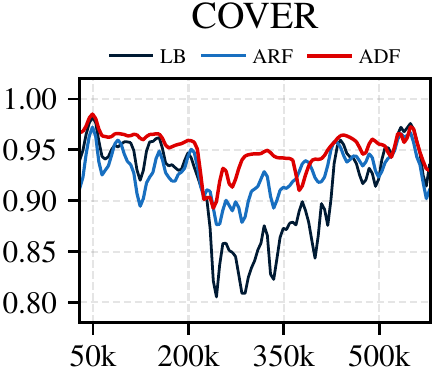}
		\end{subfigure}\\
		\vspace{3pt}
		\begin{subfigure}{0.32\linewidth}
			\centering
			\includegraphics[width=\linewidth]{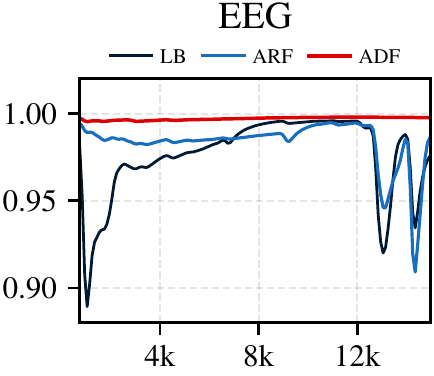}
		\end{subfigure}
		\begin{subfigure}{0.32\linewidth}
			\centering
			\includegraphics[width=\linewidth]{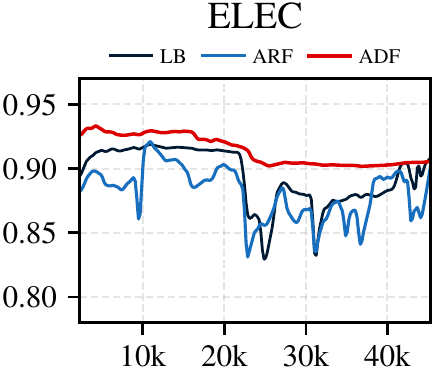}
		\end{subfigure}
		\begin{subfigure}{0.32\linewidth}
			\centering
			\includegraphics[width=\linewidth]{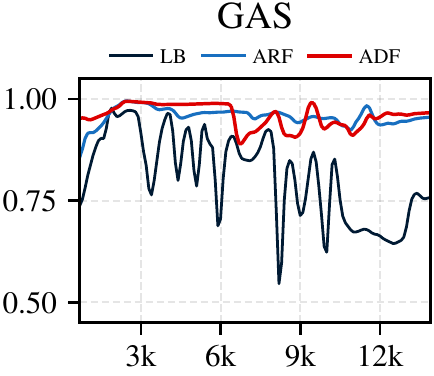}
		\end{subfigure}\\
		\vspace{3pt}
		\begin{subfigure}{0.32\linewidth}
			\centering
			\includegraphics[width=\linewidth]{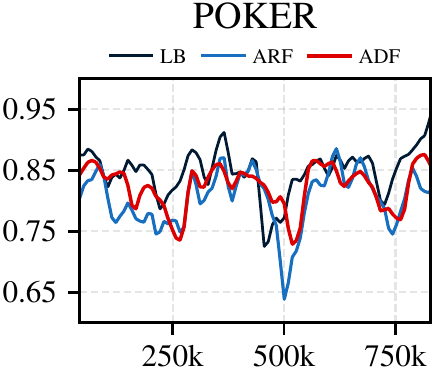}
		\end{subfigure}
		\begin{subfigure}{0.32\linewidth}
			\centering
			\includegraphics[width=\linewidth]{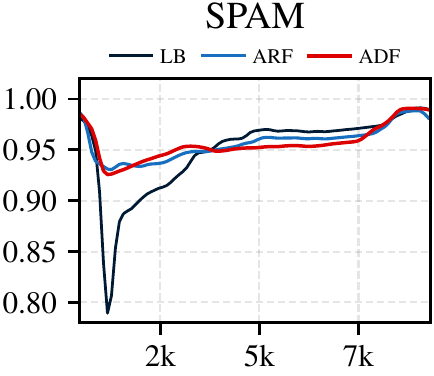}
		\end{subfigure}
		\begin{subfigure}{0.32\linewidth}
			\centering
			\includegraphics[width=\linewidth]{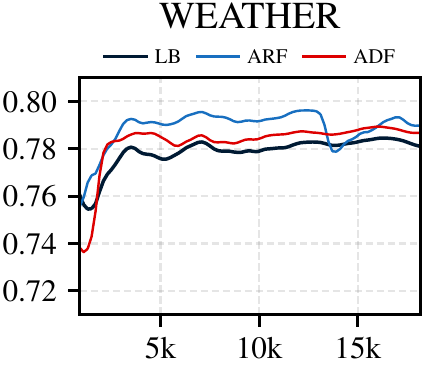}
		\end{subfigure}
		\caption{Accuracy series for ADF, ARF and LB obtained for the shallow data streams.}
		\label{fig:series-1}
	\end{figure}

\begin{figure}[htb]
		\centering
		\begin{subfigure}{0.32\linewidth}
			\centering
			\includegraphics[width=\linewidth]{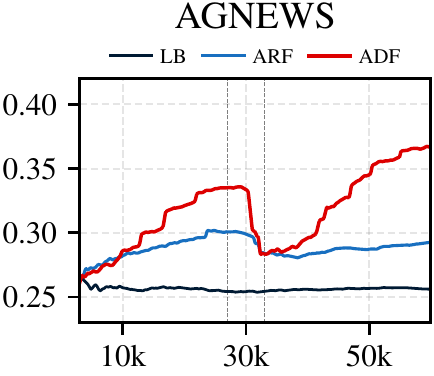}
		\end{subfigure}
		\begin{subfigure}{0.32\linewidth}
			\centering
			\includegraphics[width=\linewidth]{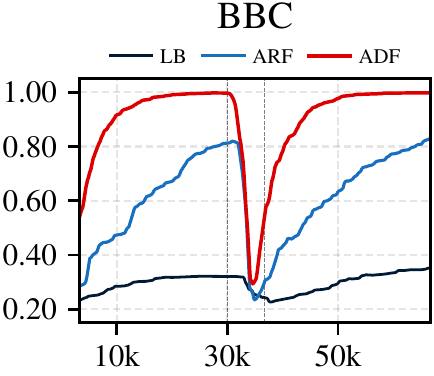}
		\end{subfigure}
		\begin{subfigure}{0.32\linewidth}
			\centering
			\includegraphics[width=\linewidth]{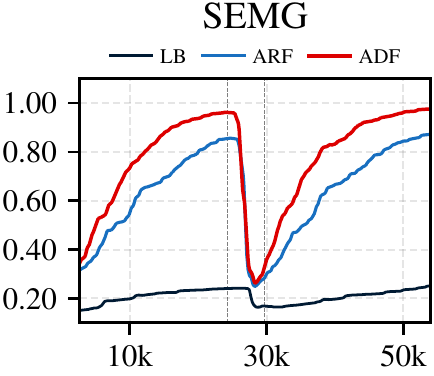}
		\end{subfigure}\\
		\vspace{3pt}
		\begin{subfigure}{0.32\linewidth}
			\centering
			\includegraphics[width=\linewidth]{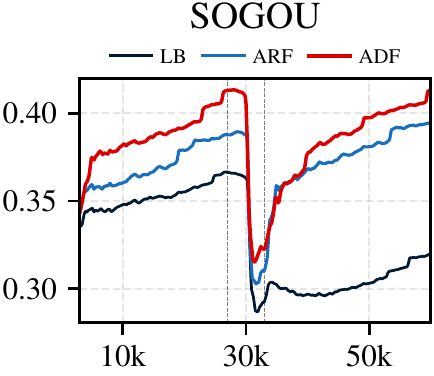}
		\end{subfigure}
		\begin{subfigure}{0.32\linewidth}
			\centering
			\includegraphics[width=\linewidth]{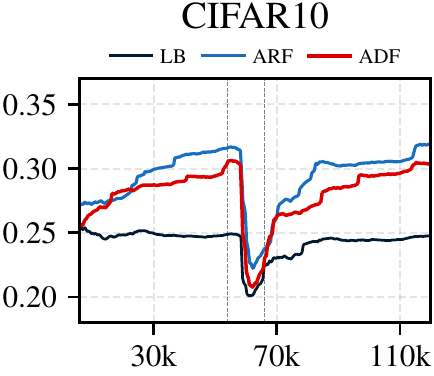}
		\end{subfigure}
		\begin{subfigure}{0.32\linewidth}
			\centering
			\includegraphics[width=\linewidth]{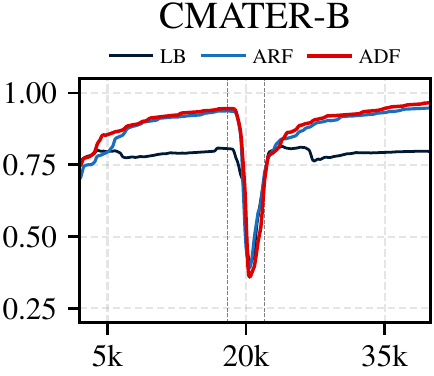}
		\end{subfigure}\\
		\vspace{3pt}
		\begin{subfigure}{0.32\linewidth}
			\centering
			\includegraphics[width=\linewidth]{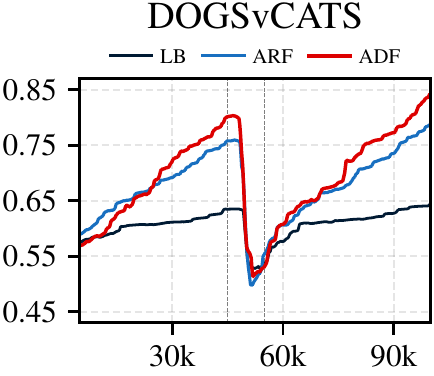}
		\end{subfigure}
		\begin{subfigure}{0.32\linewidth}
			\centering
			\includegraphics[width=\linewidth]{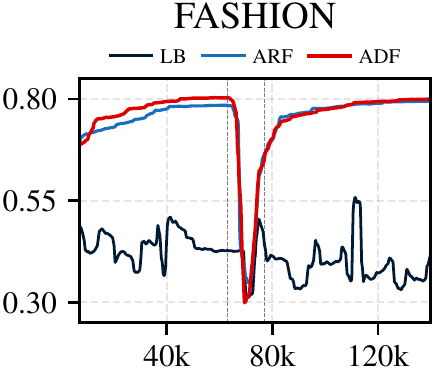}
		\end{subfigure}
		\begin{subfigure}{0.32\linewidth}
			\centering
			\includegraphics[width=\linewidth]{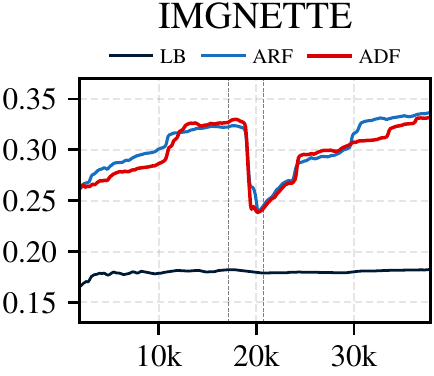}
		\end{subfigure}\\
		\vspace{3pt}
		\begin{subfigure}{0.32\linewidth}
			\centering
			\includegraphics[width=\linewidth]{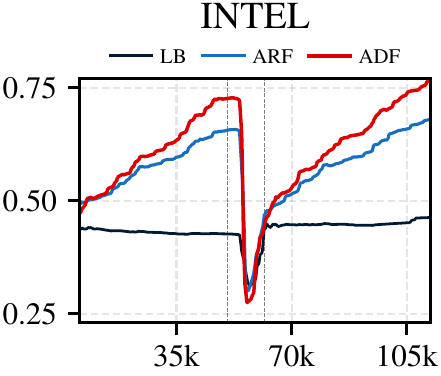}
		\end{subfigure}
		\begin{subfigure}{0.32\linewidth}
			\centering
			\includegraphics[width=\linewidth]{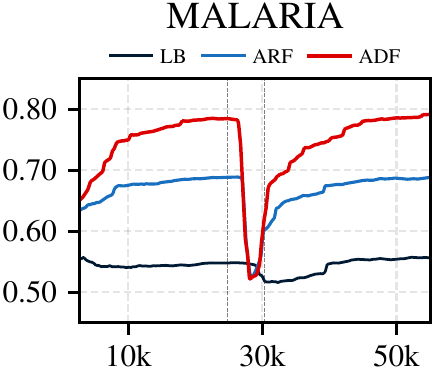}
		\end{subfigure}
		\begin{subfigure}{0.32\linewidth}
			\centering
			\includegraphics[width=\linewidth]{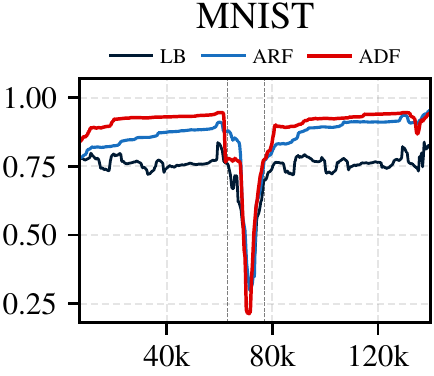}
		\end{subfigure}
		\caption{Accuracy series for ADF, ARF and LB obtained for the deep data streams.}
		\label{fig:series-2}
	\end{figure}

	In Fig. \ref{fig:series-1} and \ref{fig:series-2}, we presented the temporal performance (accuracy) of ADF compared with its baseline ARF and the second-best another ensemble -- LB. These particular examples show that ADF can efficiently handle both shallow and complex drifting streams, usually more efficiently than any other classifier. The improvements can be seen as a higher and more stable saturation (ACTIVITY, COVER, EEG, ELEC or GAS) or more accurate learning of concepts (all deep streams except for CIFAR10 and IMGNETTE). Moreover, despite of its deep structure and numerous learning units, ADF is capable of swiftly recovering from concept drifts (marked by dotted lines), which is a very important observation. The results also show how very efficient algorithm designed for shallow learning, like LB, may fail at modeling more complex concepts  (\textbf{RQ3 answered}). Finally, we can see that in many cases (AGNEWS, SOGOU, DOGSvCATS, INTEL) ADF does not reach its potential saturation points before a concept drift occurs. This suggests that there may be some room for improvements regarding the speed of convergence.
	
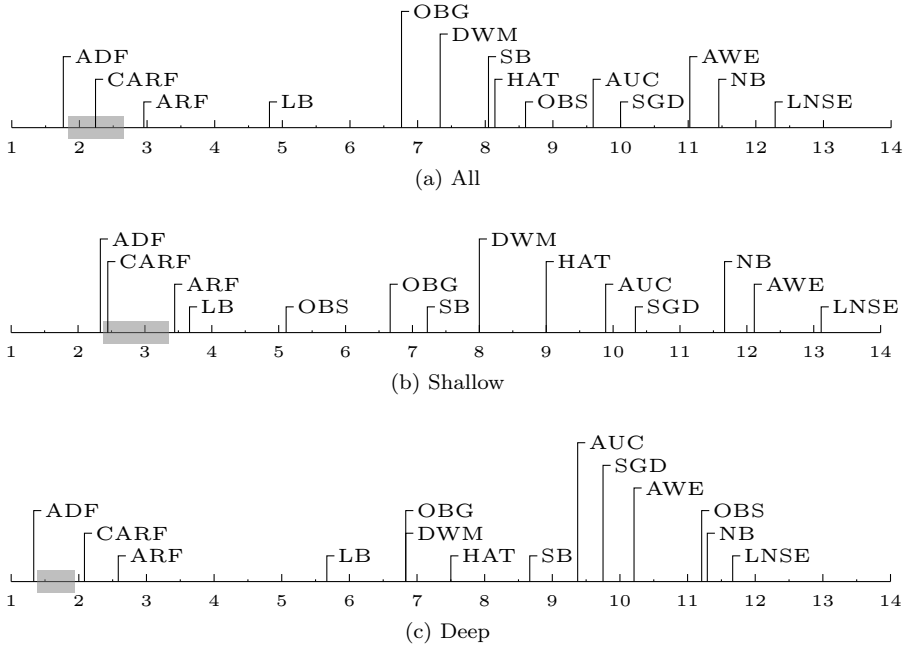
\begin{figure}[!htb]
		\centering
		\small
		{\begin{subfigure}{\columnwidth}
				\begin{tikzpicture}
				\pgfmathsetmacro{\sc}{0.89}
				\pgfmathsetmacro{\b}{1}
				
				\draw (0,0) -- ({\sc*(14-\b)},0);
				\foreach \x in {1,2,3,4,5,6,7,8,9,10,11,12,13,14} {
					\draw ({\sc*(\x-\b)}, 0) -- ++(0,.1) node [below=0.15cm,scale=1.3] {\tiny \x};
					\ifthenelse{\x < 13}{\draw ({\sc*(\x-\b) +.5*\sc}, 0) -- ++(0,.03);}{}
				}
				\coordinate (c0) at ({\sc*(1.761905-\b)},0);
				\coordinate (c1) at ({\sc*(2.238095-\b)},0);
				\coordinate (c2) at ({\sc*(2.952381-\b)},0);
				\coordinate (c3) at ({\sc*(4.809524-\b)},0);
				\coordinate (c4) at ({\sc*(6.761905-\b)},0);
				\coordinate (c5) at ({\sc*(7.333333-\b)},0);
				\coordinate (c6) at ({\sc*(8.047619-\b)},0);
				\coordinate (c7) at ({\sc*(8.142857-\b)},0);
				\coordinate (c8) at ({\sc*(8.595238-\b)},0);
				\coordinate (c9) at ({\sc*(9.595238-\b)},0);
				\coordinate (c10) at ({\sc*(10-\b)},0);
				\coordinate (c11) at ({\sc*(11.02381-\b)},0);
				\coordinate (c12) at ({\sc*(11.452381-\b)},0);
				\coordinate (c13) at ({\sc*(12.285714-\b)},0);
				
				\node (l0) at (c0) [above right=.7cm and .1cm, align=left, scale=1.45] {\hspace*{-.075cm}\tiny ADF};
				\node (l1) at (c1) [above right=.4cm and .1cm, align=center, scale=1.45] {\hspace*{-.075cm}\tiny CARF};
				\node (l2) at (c2) [above right=.1cm and .1cm, align=left, scale=1.45] {\hspace*{-.075cm}\tiny ARF};
				\node (l3) at (c3) [above right=.1cm and .1cm, align=center, scale=1.45] {\hspace*{-.075cm}\tiny LB};
				\node (l4) at (c4) [above right=1.3cm and .1cm, align=left, scale=1.45] {\hspace*{-.075cm}\tiny OBG};
				\node (l5) at (c5) [above right=1.0cm and .1cm, align=left, scale=1.45] {\hspace*{-.075cm}\tiny DWM};
				\node (l6) at (c6) [above right=.7cm and .1cm, align=center, scale=1.45] {\hspace*{-.075cm}\tiny SB};
				\node (l7) at (c7) [above right=.4cm and .1cm, align=center, scale=1.45] {\hspace*{-.075cm}\tiny HAT};
				\node (l8) at (c8) [above right=.1cm and .1cm, align=left, scale=1.45] {\hspace*{-.075cm}\tiny OBS};
				\node (l9) at (c9) [above right=.4cm and .1cm, align=left, scale=1.45] {\hspace*{-.075cm}\tiny AUC};
				\node (l10) at (c10) [above right=.1cm and .1cm, align=left, scale=1.45] {\hspace*{-.075cm}\tiny SGD};
				\node (l11) at (c11) [above right=.7cm and .1cm, align=left, scale=1.45] {\hspace*{-.075cm}\tiny AWE};
				\node (l12) at (c12) [above right=.4cm and .1cm, align=left, scale=1.45] {\hspace*{-.075cm}\tiny NB};
				\node (l13) at (c13) [above right=.1cm and .1cm, align=left, scale=1.45] {\hspace*{-.075cm}\tiny LNSE};
				
				\fill[fill=gray,fill opacity=0.5] ({\sc*(1.83-\b)},-0.15) rectangle ({\sc*(1.85+0.7963-\b)},0.15);
				
				\foreach \x in {0,...,13} {
					\draw (l\x) -| (c\x);
				};
				\end{tikzpicture}
			\end{subfigure}\vspace*{-0.15cm}\subcaption{All}}
		\vspace*{0.35cm}
		{\begin{subfigure}{\columnwidth}
				\begin{tikzpicture}
				\pgfmathsetmacro{\sc}{0.88}
				\pgfmathsetmacro{\b}{1}
				
				\draw (0,0) -- ({\sc*(14-\b)},0);
				\foreach \x in {1,2,3,4,5,6,7,8,9,10,11,12,13,14} {
					\draw ({\sc*(\x-\b)}, 0) -- ++(0,.1) node [below=0.15cm,scale=1.25] {\tiny \x};
					\ifthenelse{\x < 13}{\draw ({\sc*(\x-\b) +.5*\sc}, 0) -- ++(0,.03);}{}
				}
				\coordinate (c0) at ({\sc*(2.333333-\b)},0);
				\coordinate (c1) at ({\sc*(2.444444-\b)},0);
				\coordinate (c2) at ({\sc*(3.444444-\b)},0);
				\coordinate (c3) at ({\sc*(3.666667-\b)},0);
				\coordinate (c4) at ({\sc*(5.111111-\b)},0);
				\coordinate (c5) at ({\sc*(6.666667-\b)},0);
				\coordinate (c6) at ({\sc*(7.222222-\b)},0);
				\coordinate (c7) at ({\sc*(8-\b)},0);
				\coordinate (c8) at ({\sc*(9-\b)},0);
				\coordinate (c9) at ({\sc*(9.888889-\b)},0);
				\coordinate (c10) at ({\sc*(10.333333-\b)},0);
				\coordinate (c11) at ({\sc*(11.666667-\b)},0);
				\coordinate (c12) at ({\sc*(12.111111-\b)},0);
				\coordinate (c13) at ({\sc*(13.111111-\b)},0);
				
				\node (l0) at (c0) [above right=1.0cm and .1cm, align=left, scale=1.45] {\hspace*{-.075cm}\tiny ADF};
				\node (l1) at (c1) [above right=.7cm and .1cm, align=center, scale=1.45] {\hspace*{-.075cm}\tiny CARF};
				\node (l2) at (c2) [above right=.4cm and .1cm, align=left, scale=1.45] {\hspace*{-.075cm}\tiny ARF};
				\node (l3) at (c3) [above right=.1cm and .1cm, align=center, scale=1.45] {\hspace*{-.075cm}\tiny LB};
				\node (l4) at (c4) [above right=.1cm and .1cm, align=left, scale=1.45] {\hspace*{-.075cm}\tiny OBS};
				\node (l5) at (c5) [above right=.4cm and .1cm, align=left, scale=1.45] {\hspace*{-.075cm}\tiny OBG};
				\node (l6) at (c6) [above right=.1cm and .1cm, align=center, scale=1.45] {\hspace*{-.075cm}\tiny SB};
				\node (l7) at (c7) [above right=1.0cm and .1cm, align=center, scale=1.45] {\hspace*{-.075cm}\tiny DWM};
				\node (l8) at (c8) [above right=.7cm and .1cm, align=left, scale=1.45] {\hspace*{-.075cm}\tiny HAT};
				\node (l9) at (c9) [above right=.4cm and .1cm, align=left, scale=1.45] {\hspace*{-.075cm}\tiny AUC};
				\node (l10) at (c10) [above right=.1cm and .1cm, align=left, scale=1.45] {\hspace*{-.075cm}\tiny SGD};
				\node (l11) at (c11) [above right=.7cm and .1cm, align=left, scale=1.45] {\hspace*{-.075cm}\tiny NB};
				\node (l12) at (c12) [above right=.4cm and .1cm, align=left, scale=1.45] {\hspace*{-.075cm}\tiny AWE};
				\node (l13) at (c13) [above right=.1cm and .1cm, align=left, scale=1.45] {\hspace*{-.075cm}\tiny LNSE};
				
				\fill[fill=gray,fill opacity=0.5] ({\sc*(2.375-\b)},-0.15) rectangle ({\sc*(2.35+0.9986-\b)},0.15);
				
				\foreach \x in {0,...,13} {
					\draw (l\x) -| (c\x);
				};
				\end{tikzpicture}
			\end{subfigure}\vspace*{-0.15cm}\subcaption{Shallow}}
		\vspace*{0.35cm}
		{\begin{subfigure}{\columnwidth}
				\begin{tikzpicture}
				\pgfmathsetmacro{\sc}{0.89}
				\pgfmathsetmacro{\b}{1}
				
				\draw (0,0) -- ({\sc*(14-\b)},0);
				\foreach \x in {1,2,3,4,5,6,7,8,9,10,11,12,13,14} {
					\draw ({\sc*(\x-\b)}, 0) -- ++(0,.1) node [below=0.15cm,scale=1.25] {\tiny \x};
					\ifthenelse{\x < 14}{\draw ({\sc*(\x-\b) +.5*\sc}, 0) -- ++(0,.03);}{}
				}
				\coordinate (c0) at ({\sc*(1.333333-\b)},0);
				\coordinate (c1) at ({\sc*(2.083333-\b)},0);
				\coordinate (c2) at ({\sc*(2.583333-\b)},0);
				\coordinate (c3) at ({\sc*(5.666667-\b)},0);
				\coordinate (c4) at ({\sc*(6.833333-\b)},0);
				\coordinate (c5) at ({\sc*(6.833333-\b)},0);
				\coordinate (c6) at ({\sc*(7.5-\b)},0);
				\coordinate (c7) at ({\sc*(8.666667-\b)},0);
				\coordinate (c8) at ({\sc*(9.375-\b)},0);
				\coordinate (c9) at ({\sc*(9.75-\b)},0);
				\coordinate (c10) at ({\sc*(10.208333-\b)},0);
				\coordinate (c11) at ({\sc*(11.208333-\b)},0);
				\coordinate (c12) at ({\sc*(11.291667-\b)},0);
				\coordinate (c13) at ({\sc*(11.666667-\b)},0);
				
				\node (l0) at (c0) [above right=.7cm and .1cm, align=left, scale=1.45] {\hspace*{-.075cm}\tiny ADF};
				\node (l1) at (c1) [above right=.4cm and .1cm, align=center, scale=1.45] {\hspace*{-.075cm}\tiny CARF};
				\node (l2) at (c2) [above right=.1cm and .1cm, align=left, scale=1.45] {\hspace*{-.075cm}\tiny ARF};
				\node (l3) at (c3) [above right=.1cm and .1cm, align=center, scale=1.45] {\hspace*{-.075cm}\tiny LB};
				\node (l4) at (c4) [above right=0.7cm and .1cm, align=left, scale=1.45] {\hspace*{-.075cm}\tiny OBG};
				\node (l5) at (c5) [above right=.4cm and .1cm, align=left, scale=1.45] {\hspace*{-.075cm}\tiny DWM};
				\node (l6) at (c6) [above right=.1cm and .1cm, align=center, scale=1.45] {\hspace*{-.075cm}\tiny HAT};
				\node (l7) at (c7) [above right=.1cm and .1cm, align=center, scale=1.45] {\hspace*{-.075cm}\tiny SB};
				\node (l8) at (c8) [above right=1.6cm and .1cm, align=left, scale=1.45] {\hspace*{-.075cm}\tiny AUC};
				\node (l9) at (c9) [above right=1.3cm and .1cm, align=left, scale=1.45] {\hspace*{-.075cm}\tiny SGD};
				\node (l10) at (c10) [above right=1.0cm and .1cm, align=left, scale=1.45] {\hspace*{-.075cm}\tiny AWE};
				\node (l11) at (c11) [above right=.7cm and .1cm, align=left, scale=1.45] {\hspace*{-.075cm}\tiny OBS};
				\node (l12) at (c12) [above right=.4cm and .1cm, align=left, scale=1.45] {\hspace*{-.075cm}\tiny NB};
				\node (l13) at (c13) [above right=.1cm and .1cm, align=left, scale=1.45] {\hspace*{-.075cm}\tiny LNSE};
				
				\fill[fill=gray,fill opacity=0.5] ({\sc*(1.38-\b)},-0.15) rectangle ({\sc*(1.35+0.5891-\b)},0.15);
				
				\foreach \x in {0,...,13} {
					\draw (l\x) -| (c\x);
				};
				\end{tikzpicture}
			\end{subfigure}\vspace*{-0.15cm}\subcaption{Deep}}
		\caption{The Bonferroni-Dunn test for all classifiers and different types of streams (prequential accuracy).}
		\label{fig:bonferroni_ext_acc}
	\end{figure}

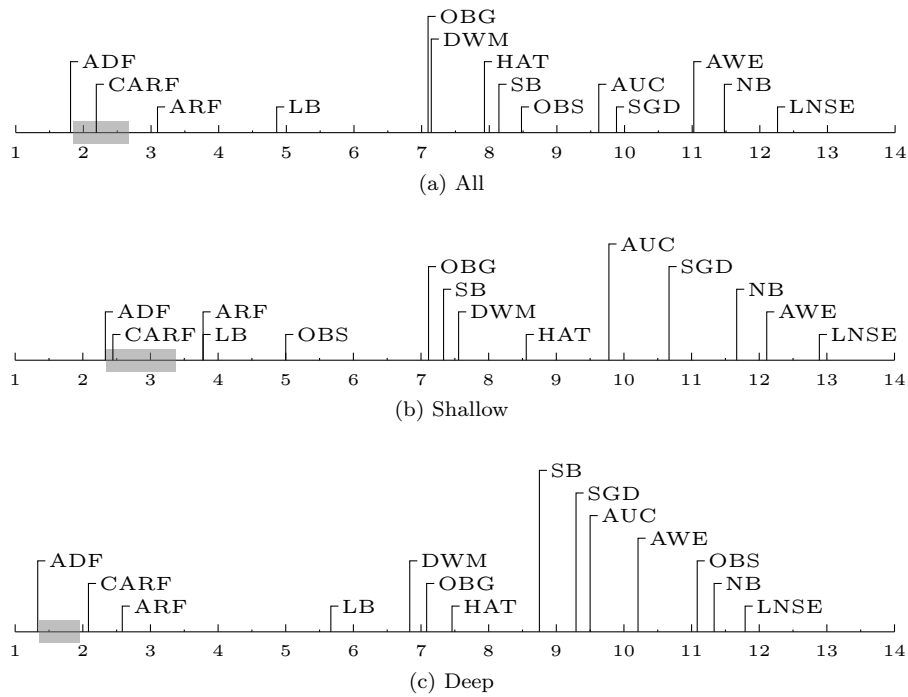
\begin{figure}[!htb]
		\centering
		\small
		{\begin{subfigure}{\columnwidth}
				\begin{tikzpicture}
				\pgfmathsetmacro{\sc}{0.89}
				\pgfmathsetmacro{\b}{1}
				
				\draw (0,0) -- ({\sc*(14-\b)},0);
				\foreach \x in {1,2,3,4,5,6,7,8,9,10,11,12,13,14} {
					\draw ({\sc*(\x-\b)}, 0) -- ++(0,.1) node [below=0.15cm,scale=1.3] {\tiny \x};
					\ifthenelse{\x < 13}{\draw ({\sc*(\x-\b) +.5*\sc}, 0) -- ++(0,.03);}{}
				}
				\coordinate (c0) at ({\sc*(1.8095-\b)},0);
				\coordinate (c1) at ({\sc*(2.1905-\b)},0);
				\coordinate (c2) at ({\sc*(3.0952-\b)},0);
				\coordinate (c3) at ({\sc*(4.8571-\b)},0);
				\coordinate (c4) at ({\sc*(7.0952-\b)},0);
				\coordinate (c5) at ({\sc*(7.1429-\b)},0);
				\coordinate (c6) at ({\sc*(7.9286-\b)},0);
				\coordinate (c7) at ({\sc*(8.1429-\b)},0);
				\coordinate (c8) at ({\sc*(8.4762-\b)},0);
				\coordinate (c9) at ({\sc*(9.619-\b)},0);
				\coordinate (c10) at ({\sc*(9.881-\b)},0);
				\coordinate (c11) at ({\sc*(11.0238-\b)},0);
				\coordinate (c12) at ({\sc*(11.4762-\b)},0);
				\coordinate (c13) at ({\sc*(12.2619-\b)},0);
				
				\node (l0) at (c0) [above right=.7cm and .1cm, align=left, scale=1.45] {\hspace*{-.075cm}\tiny ADF};
				\node (l1) at (c1) [above right=.4cm and .1cm, align=center, scale=1.45] {\hspace*{-.075cm}\tiny CARF};
				\node (l2) at (c2) [above right=.1cm and .1cm, align=left, scale=1.45] {\hspace*{-.075cm}\tiny ARF};
				\node (l3) at (c3) [above right=.1cm and .1cm, align=center, scale=1.45] {\hspace*{-.075cm}\tiny LB};
				\node (l4) at (c4) [above right=1.3cm and .1cm, align=left, scale=1.45] {\hspace*{-.075cm}\tiny OBG};
				\node (l5) at (c5) [above right=1.0cm and .1cm, align=left, scale=1.45] {\hspace*{-.075cm}\tiny DWM};
				\node (l6) at (c6) [above right=.7cm and .1cm, align=center, scale=1.45] {\hspace*{-.075cm}\tiny HAT};
				\node (l7) at (c7) [above right=.4cm and .1cm, align=center, scale=1.45] {\hspace*{-.075cm}\tiny SB};
				\node (l8) at (c8) [above right=.1cm and .1cm, align=left, scale=1.45] {\hspace*{-.075cm}\tiny OBS};
				\node (l9) at (c9) [above right=.4cm and .1cm, align=left, scale=1.45] {\hspace*{-.075cm}\tiny AUC};
				\node (l10) at (c10) [above right=.1cm and .1cm, align=left, scale=1.45] {\hspace*{-.075cm}\tiny SGD};
				\node (l11) at (c11) [above right=.7cm and .1cm, align=left, scale=1.45] {\hspace*{-.075cm}\tiny AWE};
				\node (l12) at (c12) [above right=.4cm and .1cm, align=left, scale=1.45] {\hspace*{-.075cm}\tiny NB};
				\node (l13) at (c13) [above right=.1cm and .1cm, align=left, scale=1.45] {\hspace*{-.075cm}\tiny LNSE};
				
				\fill[fill=gray,fill opacity=0.5] ({\sc*(1.85-\b)},-0.15) rectangle ({\sc*(1.85+0.8274-\b)},0.15);
				
				\foreach \x in {0,...,13} {
					\draw (l\x) -| (c\x);
				};
				\end{tikzpicture}
			\end{subfigure}\vspace*{-0.15cm}\subcaption{All}}
		\vspace*{0.35cm}
		{\begin{subfigure}{\columnwidth}
				\begin{tikzpicture}
				\pgfmathsetmacro{\sc}{0.89}
				\pgfmathsetmacro{\b}{1}
				
				\draw (0,0) -- ({\sc*(14-\b)},0);
				\foreach \x in {1,2,3,4,5,6,7,8,9,10,11,12,13,14} {
					\draw ({\sc*(\x-\b)}, 0) -- ++(0,.1) node [below=0.15cm,scale=1.25] {\tiny \x};
					\ifthenelse{\x < 13}{\draw ({\sc*(\x-\b) +.5*\sc}, 0) -- ++(0,.03);}{}
				}
				\coordinate (c0) at ({\sc*(2.3333-\b)},0);
				\coordinate (c1) at ({\sc*(2.4444-\b)},0);
				\coordinate (c2) at ({\sc*(3.7778-\b)},0);
				\coordinate (c3) at ({\sc*(3.7778-\b)},0);
				\coordinate (c4) at ({\sc*(5-\b)},0);
				\coordinate (c5) at ({\sc*(7.1111-\b)},0);
				\coordinate (c6) at ({\sc*(7.3333-\b)},0);
				\coordinate (c7) at ({\sc*(7.5556-\b)},0);
				\coordinate (c8) at ({\sc*(8.5556-\b)},0);
				\coordinate (c9) at ({\sc*(9.7778-\b)},0);
				\coordinate (c10) at ({\sc*(10.6667-\b)},0);
				\coordinate (c11) at ({\sc*(11.6667-\b)},0);
				\coordinate (c12) at ({\sc*(12.1111-\b)},0);
				\coordinate (c13) at ({\sc*(12.8889-\b)},0);
				
				\node (l0) at (c0) [above right=.4cm and .1cm, align=left, scale=1.45] {\hspace*{-.075cm}\tiny ADF};
				\node (l1) at (c1) [above right=.1cm and .1cm, align=center, scale=1.45] {\hspace*{-.075cm}\tiny CARF};
				\node (l2) at (c2) [above right=.4cm and .1cm, align=left, scale=1.45] {\hspace*{-.075cm}\tiny ARF};
				\node (l3) at (c3) [above right=.1cm and .1cm, align=center, scale=1.45] {\hspace*{-.075cm}\tiny LB};
				\node (l4) at (c4) [above right=.1cm and .1cm, align=left, scale=1.45] {\hspace*{-.075cm}\tiny OBS};
				\node (l5) at (c5) [above right=1.0cm and .1cm, align=left, scale=1.45] {\hspace*{-.075cm}\tiny OBG};
				\node (l6) at (c6) [above right=.7cm and .1cm, align=center, scale=1.45] {\hspace*{-.075cm}\tiny SB};
				\node (l7) at (c7) [above right=.4cm and .1cm, align=center, scale=1.45] {\hspace*{-.075cm}\tiny DWM};
				\node (l8) at (c8) [above right=.1cm and .1cm, align=left, scale=1.45] {\hspace*{-.075cm}\tiny HAT};
				\node (l9) at (c9) [above right=1.3cm and .1cm, align=left, scale=1.45] {\hspace*{-.075cm}\tiny AUC};
				\node (l10) at (c10) [above right=1.0cm and .1cm, align=left, scale=1.45] {\hspace*{-.075cm}\tiny SGD};
				\node (l11) at (c11) [above right=.7cm and .1cm, align=left, scale=1.45] {\hspace*{-.075cm}\tiny NB};
				\node (l12) at (c12) [above right=.4cm and .1cm, align=left, scale=1.45] {\hspace*{-.075cm}\tiny AWE};
				\node (l13) at (c13) [above right=.1cm and .1cm, align=left, scale=1.45] {\hspace*{-.075cm}\tiny LNSE};
				
				\fill[fill=gray,fill opacity=0.5] ({\sc*(2.35-\b)},-0.15) rectangle ({\sc*(2.35+1.0259-\b)},0.15);
				
				\foreach \x in {0,...,13} {
					\draw (l\x) -| (c\x);
				};
				\end{tikzpicture}
			\end{subfigure}\vspace*{-0.15cm}\subcaption{Shallow}}
		\vspace*{0.35cm}
		{\begin{subfigure}{\columnwidth}
				\begin{tikzpicture}
				\pgfmathsetmacro{\sc}{0.89}
				\pgfmathsetmacro{\b}{1}
				
				\draw (0,0) -- ({\sc*(14-\b)},0);
				\foreach \x in {1,2,3,4,5,6,7,8,9,10,11,12,13,14} {
					\draw ({\sc*(\x-\b)}, 0) -- ++(0,.1) node [below=0.15cm,scale=1.25] {\tiny \x};
					\ifthenelse{\x < 14}{\draw ({\sc*(\x-\b) +.5*\sc}, 0) -- ++(0,.03);}{}
				}
				\coordinate (c0) at ({\sc*(1.3333-\b)},0);
				\coordinate (c1) at ({\sc*(2.0833-\b)},0);
				\coordinate (c2) at ({\sc*(2.5833-\b)},0);
				\coordinate (c3) at ({\sc*(5.6667-\b)},0);
				\coordinate (c4) at ({\sc*(6.8333-\b)},0);
				\coordinate (c5) at ({\sc*(7.0833-\b)},0);
				\coordinate (c6) at ({\sc*(7.4583-\b)},0);
				\coordinate (c7) at ({\sc*(8.75-\b)},0);
				\coordinate (c8) at ({\sc*(9.2917-\b)},0);
				\coordinate (c9) at ({\sc*(9.5-\b)},0);
				\coordinate (c10) at ({\sc*(10.2083-\b)},0);
				\coordinate (c11) at ({\sc*(11.0833-\b)},0);
				\coordinate (c12) at ({\sc*(11.3333-\b)},0);
				\coordinate (c13) at ({\sc*(11.7917-\b)},0);
				
				\node (l0) at (c0) [above right=.7cm and .1cm, align=left, scale=1.45] {\hspace*{-.075cm}\tiny ADF};
				\node (l1) at (c1) [above right=.4cm and .1cm, align=center, scale=1.45] {\hspace*{-.075cm}\tiny CARF};
				\node (l2) at (c2) [above right=.1cm and .1cm, align=left, scale=1.45] {\hspace*{-.075cm}\tiny ARF};
				\node (l3) at (c3) [above right=.1cm and .1cm, align=center, scale=1.45] {\hspace*{-.075cm}\tiny LB};
				\node (l4) at (c4) [above right=0.7cm and .1cm, align=left, scale=1.45] {\hspace*{-.075cm}\tiny DWM};
				\node (l5) at (c5) [above right=.4cm and .1cm, align=left, scale=1.45] {\hspace*{-.075cm}\tiny OBG};
				\node (l6) at (c6) [above right=.1cm and .1cm, align=center, scale=1.45] {\hspace*{-.075cm}\tiny HAT};
				\node (l7) at (c7) [above right=1.9cm and .1cm, align=center, scale=1.45] {\hspace*{-.075cm}\tiny SB};
				\node (l8) at (c8) [above right=1.6cm and .1cm, align=left, scale=1.45] {\hspace*{-.075cm}\tiny SGD};
				\node (l9) at (c9) [above right=1.3cm and .1cm, align=left, scale=1.45] {\hspace*{-.075cm}\tiny AUC};
				\node (l10) at (c10) [above right=1.0cm and .1cm, align=left, scale=1.45] {\hspace*{-.075cm}\tiny AWE};
				\node (l11) at (c11) [above right=.7cm and .1cm, align=left, scale=1.45] {\hspace*{-.075cm}\tiny OBS};
				\node (l12) at (c12) [above right=.4cm and .1cm, align=left, scale=1.45] {\hspace*{-.075cm}\tiny NB};
				\node (l13) at (c13) [above right=.1cm and .1cm, align=left, scale=1.45] {\hspace*{-.075cm}\tiny LNSE};
				
				\fill[fill=gray,fill opacity=0.5] ({\sc*(1.35-\b)},-0.15) rectangle ({\sc*(1.35+0.6013-\b)},0.15);
				
				\foreach \x in {0,...,13} {
					\draw (l\x) -| (c\x);
				};
				\end{tikzpicture}
			\end{subfigure}\vspace*{-0.15cm}\subcaption{Deep}}
		\caption{The Bonferroni-Dunn test for all classifiers and different types of streams (kappa).}
		\label{fig:bonferroni_ext_kappa}
	\end{figure}

	The statistical analysis of the results, presented in Fig. \ref{fig:bonferroni_ext_acc} and \ref{fig:bonferroni_ext_kappa}, proves that most of the differences are significant, especially for deep streams. One should also notice that while for the shallow ones and for all data ADF provides a significant improvement over ARF but not over its the CARF cascade, it is significantly better than both of them for the contextual data. It can be seen especially for such streams like BBC, SEMG, DOGSvCATS, MALARIA or MNIST. This proves that the alternative approach for contextual learning based on multi-grained scanning, proposed in gcForest, is capable of capturing some significant local information about analyzed instances. On the other hand, relatively poor results for such streams like CIFAR10 (0.2891 and 0.2101) or IMGNETTE (0.3116 and 0.2345) show that there is still a lot that should be improved. The performance for similar data like INTEL (0.6234 and 0.5478) encourages us to believe that the algorithm is capable of learning such complicated patterns, but it is limited by its configuration or imperfect design. The potential modifications mentioned in the previous paragraphs may offer a solution to those problems.
	
	\smallskip

	\smallskip
	
	\begin{table}[htb]
		\caption{Average runtime [ms] for all classifiers and data streams.}
		\centering
		\scalebox{0.95}{\begin{tabular}[]{>{\arraybackslash} m{1.7cm} >{\centering\arraybackslash} m{0.9cm} >{\centering\arraybackslash} m{0.9cm} >{\centering\arraybackslash} m{0.9cm} >{\centering\arraybackslash} m{0.9cm} >{\centering\arraybackslash} m{0.9cm} >{\centering\arraybackslash} m{0.9cm} >{\centering\arraybackslash} m{0.9cm} }
			\toprule	
			Operation & HAT & SGD & NB & LB & OBG & OBS & SB\\
			\midrule
			Update & 0.78 & 0.03 & 0.02 & 8.14 & 5.55 & 6.1 & 4.7 \\
			Prediction & 0.29 & 0.01 & 0.19 & 2.81 & 3.23 & 0.22 & 0.9\\
			Total & 1.06 & 0.04 & 0.21 & 10.95 & 8.78 & 6.32 & 5.61 \\
			\bottomrule
		\end{tabular}}
		
		\vspace{0.25cm}
		
		\scalebox{0.95}{\begin{tabular}[]{>{\arraybackslash} m{1.7cm} >{\centering\arraybackslash} m{0.9cm} >{\centering\arraybackslash} m{0.9cm} >{\centering\arraybackslash} m{0.9cm} >{\centering\arraybackslash} m{0.9cm} >{\centering\arraybackslash} m{0.9cm} >{\centering\arraybackslash} m{0.9cm} >{\centering\arraybackslash} m{0.9cm} }
			\toprule	
			Operation & DWM & AUC & AWE & LNSE & ARF & CARF & ADF\\
			\midrule
			Update & 27.79 & 7.08 & 10.9 & 20.34 & 2.33 & 12.93 & 179.26\\
			Prediction & 23.15 & 0.56 & 0.9 & 1.06 & 0.38 & 2.1 & 15.37\\
			Total & 50.93 & 7.64 & 11.8 & 21.4 & 2.72 & 15.03 & 194.63\\
			\bottomrule
			\\
		\end{tabular}}
		\label{tab:time}
	\end{table}	
	
	\smallskip
	\noindent\textbf{Runtime}. The only aspect in which ADF significantly fails is, unsurprisingly, runtime. Tab. \ref{tab:time} shows the average update, prediction and total time for all streaming classifiers. It is immediately apparent that the outstanding gain in performance provided by ADF comes at cost of notable delays in processing, especially during training. ADF was almost 4 times slower (194 ms) than the second slowest DWM (51 ms) in total. However, it is worth noting that it was also faster in the prediction phase (15 ms) than this ensemble (23 ms), which suggests that the obtained runtime is not completely off base. The registered differences are not surprising as ADF uses several ensembles, while other classifiers are represented by only one of them. The runtime should be greatly reduced in more efficient implementations, since random forests can be parallelized on both CPU/GPU \citep{Marron:2014}.
	
	\begin{figure}[htb]
		\centering
		\includegraphics[width=0.7\columnwidth]{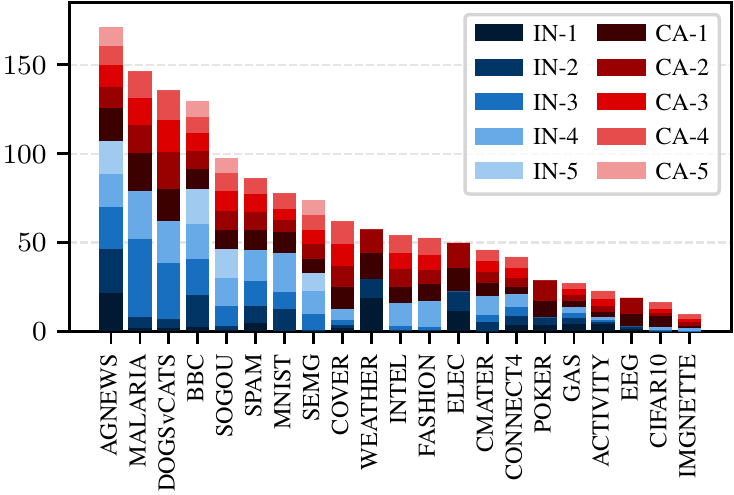}
		\caption{Average tree depth per input and cascade sublayer.}
		\label{fig:avg_tree_depth}
	\end{figure}
	
	Finally, during the experiments, we noticed that ADF notably slows down for some of the considered streams more than for others. One of them was AGNEWS, which appeared very interesting to us, since it is by no means the most complex one (only 170 attributes). Therefore, we decided to observe the average tree depth of Hoeffding trees in each sublayer, in order to visualize a relative approximation of ADF size for different streams. Fig. \ref{fig:avg_tree_depth} presents the results of these observations. Indeed, ADF produced the deepest trees for AGNEWS, which explains the distinctive total runtime we registered for this stream. In general, we can see that trees were usually almost equal for most of the cascade layers (CA), while differing in the input layer (IN). Furthermore, one should notice that while the deepest trees could be found in the input layers of the largest models (AGNEWS, MALARIA, DOGSvCATS, BBC), they were usually bigger in the cascade layer for smaller ones (COVER, INTEL, FASHION, POKER, EEG). Although a detailed analysis of the results is beyond the scope of this work, it provides some interesting insight into the relative characteristics of the data streams with respect to the capabilities of the ADF model. It is possible that for the largest models, ADF overfits to data, while for the smallest ones, especially for high-dimensional streams for which low performance was registered (CIFAR10, IMGNETTE), it fails at capturing patterns efficiently. While the former problem may possibly be handled by strictly limiting the maximum tree depth, the latter is a more complex problem, which probably may be addressed by the more advanced modifications mentioned in the previous paragraphs.
	
	\subsection{Conclusions}
	
	Based on the presented and discussed results, we can conclude with the following statements.
	
	\begin{itemize}
		\smallskip
		\item \textbf{The proposed ADF classifier is an efficient online deep forest capable of learning from non-stationary data streams.} We showed that, based on successful streaming ARF models, gcForest can be transformed into an online algorithm suitable for streaming scenarios involving incremental learning of stable concepts and adaptation to concept drifts. The complex evaluation presented in this work proves that ADF outperforms all state-of-the-art streaming algorithms, especially for high-dimensional contextual data.
		
		\smallskip
		\item \textbf{The online multi-grained scanning combined with deep cascading allows for more accurate learning from complex contextual streams.} The obtained results comparing ARF and the standalone cascade CARF suggest that ADF transfers the capabilities of the alternative deep representation learning, implemented in the original deep forest, into dynamic environments. We could observe that ADF significantly improves upon ARF and CARF when learning from visual, textual and time series data, which suggests that the online multi-grained scanning combined with representation learning in subsequent sublayers may offer some reasonable alternative approach to deep learning in online scenarios.
		
		\smallskip
		\item \textbf{There are some bottlenecks in the design and implementation that can and should be addressed.} Finally, the evaluation of different configurations and the runtime showed that the current design of ADF meets some critical limitations in these areas. In the first case, we noticed that increasing the sizes of ADF does not contribute to significant improvements in predictive performance, while in the second one, we observed critical increases in computation time for more precise and numerous models. We assume that the former can be addressed by increasing diversification of learning units, while the latter problem should be alleviated by a parallel implementation. Last bu not least, it is possible that the speed of convergence could also be improved.
		
	\end{itemize}
	
	\section{Summary}
	\label{sec:sum}

	In this work, we introduced the ADF classifier -- an online adaptive version of the deep forest algorithm, built upon the very successful ARF ensemble and capable of learning from data streams. It implements the alternative approach to handling complex contextual data, based on multi-grained scanning, representation learning and cascading, proposed for gcForest and transferred to non-stationary scenarios. 
	
	The proposed ADF framework offers a first step towards creating deep learning architectures capable of online and continual learning from streaming data, while being capable of handling the presence of concept drift. Our proposal of using adaptive decision trees as base components of a deep architecture allowed for combining the advantages of these two worlds. ADF can effectively learn from complex and high-dimensional data streams while offering excellent reactivity to any undergoing changes.  
	
	The conducted experiments show that ADF offers a much more efficient learning procedure for images, text and time series than all state-of-the-art streaming algorithms, which are limited to shallow adaptation. In the future works, we plan to scale the algorithm up by improving its utilization of numerous learning units and enabling parallel computing.
	
	\bibliographystyle{spbasic}       
	\bibliography{references}
	
\end{document}